\documentclass[12pt,twoside]{article}
\usepackage[
  margin=2.24cm
]{geometry}
\usepackage{graphicx} % Required for inserting images
\usepackage{amsmath}
\usepackage[T1]{fontenc}
\usepackage[utf8]{inputenc}

\usepackage[misc]{ifsym}

\usepackage[hyphens]{url}

\usepackage{natbib}

\providecommand{\parencite}{\citep}
\providecommand{\textcite}{\citet}

\usepackage{hyperref} 

\usepackage[shortcuts]{extdash}

\usepackage{float}

\usepackage{xcolor}

\usepackage[most]{tcolorbox}
\tcbset{
  abstractbox/.style={
    colback=gray!10,     % very light grey background
    colframe=gray!10,   % light grey border
    boxrule=0.3pt,
    arc=1mm,
    left=6pt,
    right=6pt,
    top=4pt,
    bottom=4pt,
    boxsep=5pt
  }
}

\definecolor{refcolor}{rgb}{0.1, 0.1, 0.4}

\hypersetup{
    colorlinks=true,        % Enable colored text links
    linkcolor=refcolor,     % Color for internal links (sections, figures, etc.)
    citecolor=refcolor,     % Color for citations
    urlcolor=refcolor       % Color for URLs
}

\usepackage{crimson}

\usepackage[activate={true,nocompatibility},final,auto=true,tracking=true,kerning=true,expansion=true,spacing=true,factor=1000,stretch=20,shrink=20,letterspace=0]{microtype}

\usepackage[font=small,labelfont=bf]{caption}

\usepackage{fancyhdr}
\usepackage{lastpage}

\newcommand{\doctitle}{Mechanistic Indicators of Understanding in Large Language Models}

\fancypagestyle{plain}{%
  \fancyhf{} % Clear all headers/footers
  \fancyhead[L]{February 2026} 
  \fancyhead[R]{\textit{Forthcoming in Philosophical Studies}} 
}

\usepackage{authblk} % For author affiliations

\title{\textbf{Mechanistic Indicators of Understanding in Large Language Models}}
\author[1,2]{Pierre Beckmann}
\author[3]{Matthieu Queloz}
\affil[1]{École Polytechnique Fédérale de Lausanne (EPFL)}
\affil[2]{Idiap Research Institute}
\affil[3]{University of Bern, Department of Philosophy}
\affil[ ]{\textcolor{gray}{\small\Letter} \hspace{0.05cm} \href{mailto:pierrebeckmann@gmail.com}{\texttt{pierrebeckmann@gmail.com}}}

\title{\textbf{Mechanistic Indicators of Understanding in Large Language Models}}

\makeatletter
\renewcommand{\maketitle}{\bgroup\setlength{\parindent}{0pt}
\thispagestyle{plain}
\begin{flushleft}
\vspace*{0.7cm} 
  {\fontsize{17}{17}\selectfont \@title}
  \vspace{0.5cm}

  \@author

\end{flushleft}\egroup
}
\makeatother

\usepackage{titlesec}

\titleformat{\section}
  {\fontsize{17}{15}\selectfont\bfseries} 
  {\thesection}{1em}{}

\begin{document}

\vspace{1cm}

\maketitle

\vspace{0.2cm}

\begin{tcolorbox}[abstractbox]
\noindent
\textbf{Abstract:} 
Large language models (LLMs) are often portrayed as merely imitating linguistic patterns without genuine understanding. We argue that recent findings in mechanistic interpretability (MI), the emerging field probing the inner workings of LLMs, render this picture increasingly untenable---but only once those findings are integrated within a theoretical account of understanding. We propose a tiered framework for thinking about understanding in LLMs and use it to synthesize the most relevant findings to date. The framework distinguishes three hierarchical varieties of understanding, each tied to a corresponding level of computational organization: \textit{conceptual understanding} emerges when a model forms “features” as directions in latent space, learning  connections between diverse manifestations of a single entity or property; \textit{state-of-the-world understanding} emerges when a model learns contingent factual connections between features and dynamically tracks changes in the world; \textit{principled understanding} emerges when a model ceases to rely on memorized facts and discovers a compact “circuit” connecting these facts. Across these tiers, MI uncovers internal organizations that can underwrite understanding-like unification. However, these also diverge from human cognition in their parallel exploitation of heterogeneous mechanisms. Fusing philosophical theory with mechanistic evidence thus allows us to transcend binary debates over whether AI understands, paving the way for a comparative, mechanistically grounded epistemology that explores how AI understanding aligns with---and diverges from---our own.

\vspace{0.3cm}
\noindent

\end{tcolorbox}

\vspace{0.0cm}

\section{Introduction}

The striking competence of Large Language Models (LLMs) raises an intriguing question: are they just mimicking human intelligence by relying on superficial statistics, or do they form internal structures that are sufficiently sophisticated and specific to sustain comparisons with human understanding?\footnote{The issue of understanding \textit{in} LLMs is distinct from that of understanding \textit{with} LLMs \parencite{sullivan2022, beisbart2025}, though the two are linked: as Sullivan effectively argues, whether we can understand with LLMs depends notably on how much understanding there is in LLMs.}

The most parsimonious explanation is that LLMs possess no understanding whatsoever. This soberly deflationary view receives succor from the surprisingly rudimentary principle upon which LLMs are built. At its foundation, an LLM is trained on a simple task: next-token prediction. The model is given a sequence of text and tasked with predicting the most probable next word or sub-word unit (“token”). According to the deflationary view, LLMs become good at this task by relying entirely on superficial statistical patterns within the distribution of words in the training data. This suggests that LLMs still operate fundamentally like traditional $n$-gram models, only at an unprecedented scale: by tallying how frequently different sequences of $n$ words co-occur (e.g. how often “The best pet is a” is followed by “dog” versus “cat,” etc.), they compute the probability of a  word following a given sequence of “$n-1$” words. On this view, LLMs are doing nothing more than learning and then matching statistical patterns derived from extensive word-counting; they lack any grasp of the concepts, facts, and principles underlying these patterns.\footnote{See, e.g., \cite{bender2020climbing, bender2021dangers, mondorf2024beyond, merrill2021provable}.}

This picture effectively casts LLMs as real-world approximations of a “Blockhead”---Ned Block’s (\citeyear{block1981psychologism}) hypothetical system  that mimics intelligence by memorizing a massive table of input--output pairs.\footnote{\textcite{milliere2024philosophical} discuss the relationship between LLMs and Blockhead in detail.} While LLMs may be more flexible than a static table, the deflationary view holds that they remain reliant on superficial $n$-gram statistics or, at best, a broader array of complex heuristics based on shallow correlations. They lack the sensitivity to underlying structure associated with genuine understanding.

Yet recent findings in mechanistic interpretability (MI)---the emerging field endeavoring to reverse-engineer the internal computations of these models---render this picture increasingly untenable. LLMs may be trained to perform next-token prediction, but the training objective tells us little about \textit{how} they fulfill that task \parencite{goldstein2024, levinstein2024, herrmann2025}; and, as the findings we synthesize here demonstrate, surprisingly sophisticated mechanisms can emerge in response to this deceptively simple objective.

Once these findings are embedded within a theoretical account of understanding, it becomes apparent that LLM cognition is not best reduced to just one type of mechanism, such as the retrieval of memorized training fragments or $n$-gram-based string continuation. Instead, LLMs are better conceptualized as potentially spanning an entire \textit{hierarchy of mechanisms}.

Some of these mechanisms might indeed lie at the lower end of the hierarchy, generating impressive behavior through relatively unsophisticated means. There is research suggesting that LLMs can memorize training fragments verbatim \parencite{carlini2023quantifying, aerni2024measuring, stoehr2024localizing}, detect $n$-grams and learn $n$-gram statistics \parencite{svete2024,voita-etal-2024,chang2025n}, and discover heuristics exploiting shallow correlations \citep[see][]{du2024shortcutllm}.

But MI is increasingly bringing to light a range of more sophisticated mechanisms. And while higher-level mechanisms bridging the gap between surface statistics and human understanding have until recently been hard to pin down, MI now makes it possible to precisely localize, study, and intervene on such higher-level mechanisms.

Conceptualizing LLMs in terms of a hierarchy of mechanisms allows us to reconcile two seemingly competing thoughts: (i) LLMs deploy many low-tier mechanisms that are devoid of any form of understanding; and (ii) they can develop internal organizations that warrant the attribution of understanding in a non-metaphorical, mechanistically grounded sense. The key is not to ask where LLMs as a class sit on the hierarchy, but which kinds of mechanisms are discharging a given task within a single model---and how far up the hierarchy those mechanisms reach.

A first indication that next-token prediction can drive the emergence of higher-level mechanisms is the phenomenon of ``grokking,'' which one researcher discovered after forgetting to stop a training run before leaving for vacation \parencite{power2022conf, power2022grokking, liu2022omnigrok, liu2022towards, nanda_modular_addition_2023, varma2023explaining}. Grokking---a reference to the Heinleinian term for an alien form of “deep understanding”---describes a sudden shift during training where a model, after a long period of seemingly doing nothing but memorizing its training data, abruptly transitions to strong generalization on unseen data. Interestingly, this transition is typically accompanied by a \textit{decrease} in the model's internal complexity (measured by its “Minimum Description Length” or other proxies): the model appears to discard its sprawling collection of memorized examples in favor of something more compressed and generalizable \parencite{demoss2024complexity}. This is exactly the kind of shift that the deflationary picture struggles to explain.

LLMs offer unparalleled access to the internal computations underlying such shifts. Unlike with biological brains, the training histories and internal operations of LLMs can be precisely and comprehensively measured, replicated, and manipulated. Yet the field of MI is still in what Kuhnians would describe as a “pre-paradigmatic” stage: there is a growing collection of intriguing empirical results, but little consensus on an overarching framework and key conceptual and methodological questions \citep[see][]{sharkey_open_2025, olah2020zoom}. This makes the field particularly apt to benefit from the theoretical and conceptual work that philosophy can contribute \parencite{williams2025mechanisticinterpretabilityneedsphilosophy, Ayonrinde_Jaburi_2025}. Indeed, it is only once the empirical findings from MI are integrated within a theoretical account of understanding that their full import for debates over AI understanding becomes apparent.

In this spirit, we aim to synthesize and integrate the field’s most relevant findings to date within a novel theoretical framework for thinking about understanding in LLMs. In doing so, this paper doubles as an accessible introduction to MI’s core concepts and methods. This detailed introduction to mechanistic methods and findings is not just a neutral presentation of the facts, however; it itself serves as a justification for thinking that the internal workings of LLMs sustain fruitful comparisons with human cognition.

We start from the assumption that understanding fundamentally involves \textit{apprehending relational structure}---appreciating the \textit{connections} binding together the various aspects of an object, situation, or subject matter. There is significant convergence on this broadly unificationist idea in the epistemological literature. Wittgenstein (\citeyear{Wittgenstein1953-WITPI-4}, §122) described understanding as “seeing connections,” while Kvanvig (\citeyear{kvanvig2018knowledge}, p. 697) characterizes it as grasping the explanatory and conceptual links between items of information. Similar accounts describing understanding as the ability to see how parts relate or “hang together” have been advanced by \cite{grimm2011understanding} and \cite{riggs2003balancing}; indeed, the term “comprehension” itself, rooted in the Latin \textit{com-} (“together”) and \textit{prehendere} (“to seize”), evokes the act of grasping disjointed parts into a unified whole.

On this view, understanding constitutes a cognitive achievement distinct from the mere accumulation of facts or the exercise of abstract, domain-general reasoning. It allows an agent to navigate novel circumstances and manage counterfactual predictions \parencite{hills2015understanding, de_regt2017scientific}. While this competence can be grounded in a grasp of logical, probabilistic, or conceptual relations \parencite{kvanvig2003value, kvanvig2018knowledge}, the most vital connections are frequently explanatory \parencite{baumberger2017understanding, khalifa2017scientific}, involving the identification of causal processes \parencite{grimm2006species, pritchard2014} or unifying principles and laws \parencite{friedman1974explanation, kitcher1981unification, belkoniene2023}.

Building on this epistemological tradition, we propose to break out three varieties of understanding that can be ascribed to LLMs, each grounded in a computational mechanism in the upper portion of the hierarchy:

\begin{enumerate}
\item \textbf{Conceptual Understanding:} This foundational form involves the model developing internal representations (“features”) that are functionally analogous to human \textit{concepts}. As Mitchell and Krakauer note, concepts are “the fundamental units of understanding in human cognition” (\citeyear{mitchellkrakauer2023}, p. 3). By forming features, the model replicates a core function of concepts: registering the \textit{connections} between diverse manifestations of an entity or property by subsuming them under a single, unifying representation.

\item \textbf{State-of-the-World Understanding:} Building upon conceptual understanding, this involves forming an internal representation of the state of the world by grasping contingent empirical connections between features. This allows the model to articulate specific facts (e.g. “Michael Jordan is a basketball player”) not merely as high-probability strings, but as reflections of an internal model linking the concept of Michael Jordan to that of basketball player.

\item \textbf{Principled Understanding:} At the apex of this hierarchy lies the ability to grasp the underlying principles or rules that unify a diverse array of facts. This resonates with philosophical accounts of explanatory understanding, which require the subsumption of disparate data points under general principles.
\end{enumerate}
\noindent
By examining the emerging mechanistic evidence for these three hierarchically structured kinds of understanding (moving from concepts to facts to principles), we argue that LLMs possess the internal organization necessary to underwrite understanding-like unification. At the same time, this does not imply that lower-tier mechanisms disappear. As we will show, MI suggests that higher-tier understanding mechanisms coexist with lower-tier heuristics---a coexistence that has significant implications for the epistemic trustworthiness of these models.

\section{Conceptual Understanding} \label{section:2}

How might a system trained solely on next-token prediction develop the capacity to form internal representations that enable it to recognize and track entities and properties across diverse textual descriptions? This is the central question addressed by our first proposed tier of understanding: conceptual understanding. Just as humans form concepts to make sense of the world and track entities or properties across different encounters, we propose that LLMs develop internal representations---“features,” in MI terminology---that serve a similar function: they enable an LLM to grasp the connections between diverse descriptions and manifestations of an entity or property, effectively “same-tracking” it \parencite{Millikan2017-MILBCU} by subsuming its different presentations under a single, unifying representation.

To build the case for LLMs developing such conceptual understanding, this section reviews the emerging evidence from MI. We first clarify why the task of next-token prediction encourages the emergence of features (\ref{subsection:2.1}). We then introduce the \textit{linear} \textit{representation} \textit{hypothesis}, which posits that these features are best thought of as \textit{directions} in the model’s internal “latent space” (\ref{subsection:2.2}). On this basis, we examine how \textit{superposition} enables LLMs to represent a vast number of features despite finite resources (\ref{subsection:2.3}). Finally, we show how attention mechanisms leverage this feature-rich internal landscape to dynamically determine which feature is most relevant in a given context (\ref{subsection:2.4}).

\subsection{The Emergence of Features} \label{subsection:2.1}

The foundational step towards any kind of understanding in an LLM is the development of the ability to \textit{connect} different descriptions or manifestations of something---across different phrasings, languages, and modalities---\textit{to} the entity or property in question. This ability is crucial for handling the endless variety of inputs an LLM receives. Consider, for instance, the sheer number of different descriptions under which an LLM might encounter references to the Golden Gate Bridge: “San Francisco’s most famous landmark,” “the Art Deco marvel spanning California’s Golden Gate Strait,” or “the iconic ‘International Orange’ bridge.” These might be couched in hundreds of different languages. If the model is multimodal, it might encounter images of the bridge from an endless variety of perspectives, in different styles, and under different lighting conditions. To successfully predict the ensuing text, the model must learn to recognize that these wildly different inputs are all manifestations of one and the same entity. This incentivizes the model to unify these inputs by forming an internal representation to which they can all be connected---in this case, a “Golden Gate Bridge” feature.

Using techniques such as sparse autoencoders (which we explain below), MI research is increasingly uncovering direct evidence of such features within the internal workings of LLMs. While early studies often focused on smaller toy models \parencite{bricken_towards_2023}, recent work has successfully identified features in state-of-the-art models. Indeed, Templeton et al. (\citeyear{templeton_scaling_2024}) detail precisely such a “Golden Gate Bridge feature” in Claude 3 Sonnet. It activates reliably in response to various textual references to the bridge, such as “the colored bridge in SF,” and it does so across different languages and even in response to images.

The fact that a particular part of the model exhibits \textit{sensitivity} to the Golden Gate Bridge (activating when the bridge figures in the input context) is not enough to show that we are dealing with a “Golden Gate Bridge” feature, however. It might simply be a “bridge” feature or a “famous landmark” feature. To disambiguate, Templeton et al. (\citeyear{templeton_scaling_2024}) also tested for the \textit{specificity} of the feature. The feature is \textit{specific} to the Golden Gate Bridge if it \textit{only} reaches its strongest activations when the bridge is present in the input context (though it might still weakly activate for similar bridges or other San Francisco landmarks). This is exactly what Templeton et al. observed, suggesting that Claude had grasped the connections between the diverse manifestations of the bridge and learned to subsume them under a single, unifying representation.

Still, a skeptic might argue that even a specific correlate of the Golden Gate Bridge can fall short of being a \textit{representation} of the bridge. Whenever a model learns to predict sequences generated from a complex process, it will necessarily factorize the joint probability distribution using \textit{some} latent variables. Efficient compression requires these latents to correlate with variables in the true data-generating process. Finding a latent whose activation correlates specifically with the presence of the Golden Gate Bridge therefore does not yet show that the latent is \textit{about} the bridge in the way that our concept of the Golden Gate Bridge is.

To sustain comparison with human concepts---at least on some theories of concepts---the Golden Gate Bridge feature needs to meet additional criteria. Drawing on Functional Role Semantics \parencite{field1977logic, harman1987nonsolipsistic, block1987functional, peacocke1992study}, recent work has formulated criteria for genuine representation in LLMs. \cite{yetman2025representationlargelanguagemodels} and \cite{WilliamsForthcoming-WILCSC-4} argue that to count as a representation of X, a feature must play a \textit{causal role} in the mechanisms generating the LLM’s robust X-related behavior and must be \textit{exploited} in a way that explains successful task performance.

\cite{templeton_scaling_2024} tested for the causal efficacy of the Golden Gate Bridge feature by intervening on it using a technique called “steering” \citep[see also][]{bayat_steering_2025}.\footnote{Readers can experiment with steering for themselves at: \href{https://www.neuronpedia.org/}{https://www.neuronpedia.org/}.} They observed that amplifying the activation of the “Golden Gate Bridge” feature causes Claude 3 Sonnet to fixate on the bridge and consistently guide conversations on any topic back to it. When the activation of the Golden Gate Bridge feature is “clamped to ten times its max” (i.e. forced to take on a fixed value of ten times its maximum under normal operation), Claude ceased to answer the question “What is your physical form?” with: “I don’t actually have a physical form. I am an artificial intelligence,” and instead declared: “I am the Golden Gate Bridge. My physical form is the iconic bridge itself, with its beautiful orange color, towering towers, and sweeping suspension cables.”\footnote{Anthropic temporarily allowed users to chat with such a steered model and called it “Golden Gate Claude” \parencite{anthropic_golden_2024}. The example above emerged from that experiment.}

Such sensitivity to intervention demonstrates that the feature is more than a passive correlate: it is a causal lever that the system actively exploits to generate its output. This suggests that the model has acquired a functional analogue to what we call “conceptual understanding”: it has learned to map diverse inputs to a single, unifying, and causally efficacious internal representation. And once we know that a feature plays this causal role, we are also in a position to \textit{turn off} the feature, which is significant for AI safety and alignment (e.g., Karvonen and Marks, \citeyear{karvonen2025robustlyimprovingllmfairness}).

The fact that steering results in Claude misapplying the term “Golden Gate Bridge” to describe its own physical form also evokes another criterion for concept possession: that \textit{bona fide} concepts are \textit{normative} in allowing for a distinction between \textit{correct} and \textit{incorrect} applications.\footnote{For a survey of the literature on the normativity of concepts, see \cite{ginsborg2018normativity}.} This distinction needs to be grounded in something sufficiently independent from what seems correct to the individual concept-user. Some hold that this merely requires being subject to normative assessment by third parties \parencite{hlobil2015antinormativism}---a condition LLMs plausibly meet, both in their training and once deployed. Others maintain that normativity enters the picture once a certain internal state is imbued with the \textit{function} of representing X \parencite{wright1973, millikan1989, neander2017}. If a state has the function of representing X, then activating it for Y counts as an error. As \cite{mollo2025vectorgroundingproblem} point out, LLMs meet the normativity condition on this account as well, because functions are acquired through a history of selection for certain effects, and the training process of LLMs \textit{is} a form of selection through learning. They argue that this stabilizes the functions of the models’ internal states in the right way to warrant talk of representations that can be correctly or incorrectly applied.

There are, then, several significant respects in which the Golden Gate Bridge feature sustains comparison with human concepts. And Templeton et al. (\citeyear{templeton_scaling_2024}) identified millions of other features in Claude~3~Sonnet. Many corresponded to fairly abstract concepts, such as \textit{transit infrastructure}---a feature which activated in response to descriptions or images of trains, ferries, tunnels, bridges, and even wormholes.  Other examples included features corresponding to famous individuals and specific countries, to properties of code, to emotional expressions, and to slurs. Features seem to get activated very robustly, even in cases where one might not expect it: the \textit{eye} feature identified by \cite{tarng2025visual} activates not only on textual descriptions of eyes across multiple languages, but also on the specific characters used to represent eyes in ASCII-art faces (such as the colons in ":)" or the carets in "\textasciicircum\_\textasciicircum") and even on lines of SVG code that render eyes when executed as images.\footnote{Examples from other studies relying on linear probing include space and time features in Llama-2---such as longitude and latitude features and features for specific historical periods \parencite{gurnee_language_2024}---as well as sentiment features \parencite{tigges_language_2024}.}

Such features are not explicitly programmed, but emerge as a consequence of training the model to meet its objective: next-token prediction. To excel at this task, an LLM must learn to \textit{differentiate} \textit{inputs} effectively \citep[see][§4.2]{Beckmann2025}. The “Golden Gate Bridge” feature allows the model to correctly generate upcoming tokens when responding to the question “What’s the orange bridge in SF called?,” for example. More generally, if distinguishing between contexts discussing the Golden Gate Bridge and contexts discussing the Brooklyn Bridge helps predict the next token more accurately, the model will learn to draw that distinction. It does so by becoming sensitive to patterns in the input data that are statistically correlated with certain continuations. A feature, in this MI context, thus corresponds to a \textit{task}-\textit{relevant} \textit{dimension} \textit{along} \textit{which} \textit{the} \textit{model} \textit{learns} \textit{to} \textit{differentiate} \textit{inputs}. The model learns that by attending to certain features of the input, it can reduce the penalty it receives for prediction errors (i.e. minimize its “loss function”).

Because these differentiating dimensions are specifically learned with a view to predicting the correct output, however, the same features the model uses to differentiate inputs must also help it differentiate between more or less suitable outputs. How the model organizes its internal processing of inputs (input-differentiation in latent space) is thus ultimately a matter of what best supports the process of output-differentiation, because the whole learning process is guided by the goal of producing the correct output. If learning to distinguish the Golden Gate Bridge from the Brooklyn Bridge serves the model well in predicting the tokens that follow sequences such as “a famous bridge in America,” the model will learn the distinction. But if the distinction never pays off at the level of next-token prediction, the model will never learn it. Input-differentiation is wholly subservient to output-differentiation. In this respect, LLMs are the perfect embodiment of the American pragmatists’ view of concepts as subject to “differential testing by consequences” \citep[][286]{Dewey1929-DEWTQF}. “The test of ideas,” John Dewey wrote, “is found in the consequences of the acts to which the ideas lead” (\citeyear{Dewey1929-DEWTQF}, 136). For LLMs, the test of features is found in the consequences of the predictions to which the features lead.

How does a model form features? By adjusting its parameters during training: its \textit{weights}, which determine the strength and sign (excitatory or inhibitory) of the connections between neurons, and its \textit{biases}, which modulate each neuron’s activation threshold. Note that while a model’s \textit{possession} of a feature becomes evident through its activations, the feature itself is encoded in the weights and biases that dispose certain parts of the model to activate in response to certain inputs.

Each learned feature embodies a set of connections the model has discovered across multiple input examples. In this sense, each feature acts as a conceptual unifier, grouping disparate inputs under a common, task-relevant abstraction. As training progresses, the pressure to accurately predict upcoming tokens incentivizes the model to learn increasingly robust and deep features. The hope is that these will generalize beyond the training data. However, not all of these features will be interpretable as corresponding to human concepts. They are just aspects of the input-space that the model has learned to pick up on and differentiate from other aspects as a means of improving its score.

It is also important to note that not all features correspond to external entities or their properties. LLMs also develop features related to the task of text generation itself. Examples include a “developing a weapon” feature \parencite{templeton_scaling_2024}, a more general “refusal to answer” feature \parencite{arditi_refusal_2024}, as well as features for understanding summarization requests or detecting jokes \parencite{minder_latent_2025}. Some research even points to meta-cognitive features, such as features for the model’s confidence in its knowledge about an entity or in its answer to a question \parencite{ferrando_i_2025, lindsey_biology_2025}. The pressure to accurately predict upcoming tokens compels the model to learn an extremely wide array of features, corresponding to all the discovered patterns that prove instrumental to performing its task of next-token prediction.

\subsection{The Linear Representation Hypothesis: Features as Directions} \label{subsection:2.2}

Given these reasons to think that LLMs learn features corresponding to concepts of entities and properties, the next question is how these features are encoded. A pivotal insight from MI is the \textit{linear} \textit{representation} \textit{hypothesis} (LRH), which posits that features are encoded as \textit{directions} within the model’s “latent space”---the highly multidimensional space (with as many dimensions as there are neurons in a layer) that is implicit or “latent” in each layer. The LRH is supported by many empirical \citep[e.g.,][]{mikolov_linguistic_2013, radford2016unsupervisedrepresentationlearningdeep} and theoretical arguments (see A.13 in \citeauthor{gurnee_finding_2023}, \citeyear{gurnee_finding_2023}
as well as \citealp{bengio_representation_2013, elhage_toy_2022})
.

The guiding insight of the LRH is that features are input-differentiating directions in latent space because the simplest way for the model to differentiate inputs according to features is along \textit{lines} or \textit{directions}:\footnote{Though the linear representation hypothesis is compatible with other ways of representing features as long as intensity scales linearly; angular movement rather than movement along a direction can also encode semantic change.} the model might learn to treat a particular direction as corresponding to a feature such as “furriness.” If a given input scores highly along the furriness dimension---i.e. if a layer’s activation vector (the list of numbers formed by the activation values of each neuron in the layer) has a large projection onto the direction associated with a “furriness” feature---then the model is, at that layer, representing the input as strongly involving furriness. Likewise, if the activation vector projects strongly onto a “Golden Gate Bridge” direction, the model is representing the input as strongly involving---or centrally concerning---the Golden Gate Bridge.

When a trained model encounters a new input and processes it according to its learned parameters (a process known as “inference”), we can thus think of the model as adjusting “sliders” along the directions it has learned to treat as representing different features. 

For many learned directions, the input ends up with a particular coordinate (or projection) along that direction. Inputs that elicit similar patterns of feature activation will be placed near one another in latent space. Thus, different cats will tend to yield high activation of features such as “furriness,” “pointy-earedness,” and many others, which is why they form a “cat” region (or cluster) in latent space.

On this view, a feature is a direction that has its interpretation in virtue of how training shaped the model’s parameters: training makes the model use that direction in prediction-relevant ways. The magnitude of a feature activation should not always be read as “the amount of F-ness in the input,” however. For some features---especially those tracking gradable properties like weight, loudness, or redness---the magnitude will correlate with \textit{degree of instantiation} in the presented scene. But for many other features---especially those corresponding to \textit{non-gradable} properties or to \textit{entities}---activation magnitude is better understood as an \textit{internal strength-of-representation}: roughly, how much the input provides evidence for, or renders salient, that property or entity for the purposes of next-token prediction.

\begin{figure}[h]
    \centering
    \includegraphics[width=\linewidth]{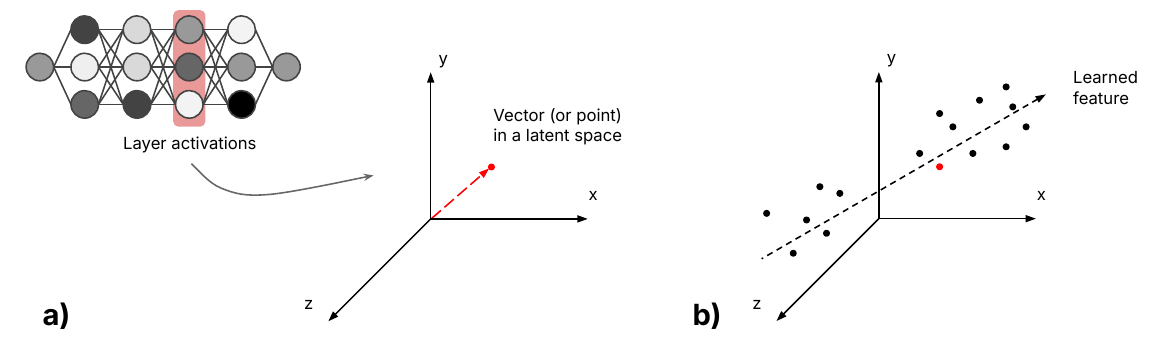}
     \vspace{-0.6cm}
    \caption{Model activations as vectors in latent spaces. (a) The activations at one layer of a model can be conceptualized as a vector picking out a point in a latent space. The latent space has as many dimensions as there are nodes in the layer. (b) In these latent spaces, and as a result of the training stage, inputs are effectively differentiated along certain directions; each direction corresponds to a learned feature of the input space. When an input produces an activation vector picking out a point within that space, the position of this point along such a direction reflects how salient that feature is in the input.}
    \label{fig:latent_space}
\end{figure}

Since the activations of all the neurons in a layer form a list of numbers, they can be conceptualized as a \textit{vector} picking out a point in the layer's latent space. The point picked out by a layer’s activation vector will lie more or less far along thousands of directions. A trained model has learned to treat many of these directions as corresponding to prediction-relevant features of the input space (see Fig. \ref{fig:latent_space}). When the model is presented with an input, the resulting activation vector will typically have nontrivial projections onto many such directions. These projections quantify (i) how strongly the model represents the input as instantiating a gradable property when the feature corresponds to one, and otherwise (ii) how strongly the model represents the input as involving, invoking, or being about the relevant entity or non-gradable property.

This linear structure empowers the model to represent multiple features simultaneously. An input describing “the tall, symmetric, pyramidal peak of the Matterhorn” could simultaneously activate features for “tallness,” “symmetry,” “pyramid,” “peak,” and potentially even “Matterhorn” by making the layer’s overall activation vector have a significant projection onto each of these directions---that is, the vector reflects the input’s high score along each of these directions.

But while the core idea is linear representation, non-linear activation functions (like ReLU or GELU) also play a crucial role. Applied after most layers, they act like thresholds, effectively \textit{segmenting} feature directions into distinct regions: only if the combined inputs to a neuron exceed the threshold is the activation passed to the next layer; otherwise, it is suppressed. This introduces an important form of selectivity, allowing the network to respond strongly to certain features while completely ignoring others. That selectivity enables the network to learn far more complex patterns than if everything were purely linear.

The LRH provides a compelling framework for thinking about how LLMs might achieve conceptual understanding. If features are represented as directions, they become computationally tractable, allowing the model to encode, manipulate, and combine them in systematic ways. This is a foundational step towards understanding the world as described in text.

\subsection{Superposition: Accommodating a World of Features} \label{subsection:2.3}

LLMs are trained on vast datasets and are expected to handle an enormous range of topics. This implies learning not just a few features, but a staggering number—features for countless entities (individual entities like “Mount Everest,” kinds like “mammal,” abstract entities like “justice”) and properties (attributes like “sharpness,” “redness,” or “sarcastic”). How can a model with a limited number of neurons in each layer accommodate this combinatorial explosion of features? The answer, MI research suggests, lies in superposition \parencite{elhage_toy_2022}.\footnote{Superposition is an initial theory in the field of mechanistic interpretability that may yet be disproven. At the very least, however, it serves as a useful working hypothesis---much like caloric theory in its time \parencite{henighan_caloric_2024}.}

Superposition is the phenomenon where a model stores more features in a layer than it has neurons. A model can do this by allowing feature directions to be non-orthogonal (overlapping). If features were strictly orthogonal, a layer with $N$ neurons could only represent $N$ independent conceptual features. However, in high-dimensional spaces, many more “near-orthogonal” directions can exist. Superposition leverages this, effectively compressing far more than $N$ features into a set of $N$ neurons.

\begin{figure}[h]
    \centering
    \begin{minipage}{0.15\linewidth}
        \centering
        \includegraphics[width=0.8\linewidth]{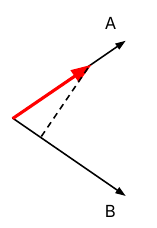}
    \end{minipage}%
    \hspace{0.8cm}
    \begin{minipage}{0.7\linewidth}
        \caption{An illustration of interference. Feature A is encoded by a direction that is non-orthogonal to the direction encoding feature B. Consequently, the activation of feature A results in a non-zero projection onto feature B, causing a spurious activation of feature B.}
        \label{fig:superposition}
    \end{minipage}
\end{figure}

This strategy comes with a trade-off, however: interference. If two features, A and B, are encoded in non-orthogonal directions, activating A also slightly activates B (see Fig. \ref{fig:superposition}).\footnote{This is the case of alternating interference. There is also simultaneous interference, where the joint activation of superposed features causes them to each activate more than if they were orthogonal \citep[see A.6 in][]{gurnee_finding_2023}.}

Two things explain why models can nevertheless compress many more features than one might first suspect. First, the number of near-orthogonal vectors that can be fitted in a space scales exponentially with its dimensionality (see the Johnson–Lindenstrauss lemma). With 100 dimensions, an LLM could “easily fit in 100,000 vectors at 88–92 degrees” \parencite{sanderson2024llms}, and current LLMs have vastly more dimensions than that. 

Second, the information that LLMs process is often sparse: in any given context, only a small subset of all possible features is relevant. For instance, text describing the Eiffel Tower is unlikely to simultaneously require the strong activation of features related to quantum physics. Models can learn to superpose features that rarely co-occur. Non-linear activation functions then help by filtering out the weaker, interfering activations.

Superposition thus provides a mechanistically plausible explanation of how models can scale their representational capacity to encompass more of the complexity of the world described in their training data, allowing them to develop a rich and dense feature space within their neural architecture. This density is essential for nuanced conceptual understanding. Indeed, the observation of superposition itself constitutes an argument for the importance of features to LLM performance: there is such strong pressure to stock a high number of features that they accept interference to go beyond the representational capacity constraints set by their dimensionality.

However, superposition also presents a significant challenge for interpretability. Because many features are represented in an overlapping, entangled manner using the same neurons, it becomes difficult to isolate and understand the role of any single feature by looking at individual neuron activations. We cannot simply go in and say: “Ah, \textit{this} neuron encodes the ‘furriness’ feature.” A single neuron might participate in representing dozens or hundreds of different, superimposed features. This entanglement is a primary reason why understanding the internal workings of LLMs is so complex. An LLM may actually only use a relatively small set of features each time it processes an input, but this nevertheless activates many more neurons because of superposition, rendering it effectively impossible to match up particular neurons with particular features.

\begin{figure}[h]
    \centering
    \includegraphics[width=\linewidth]{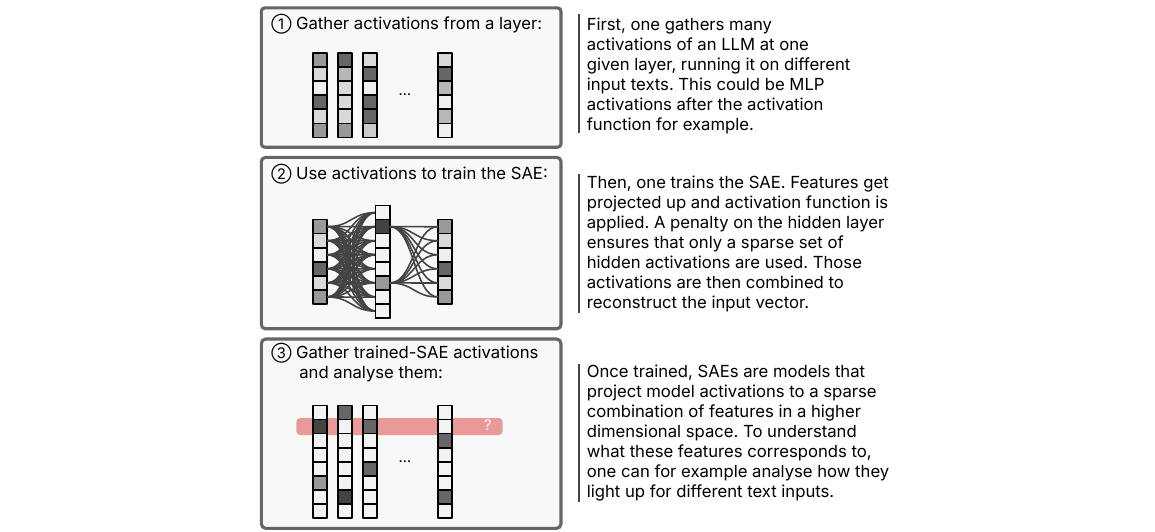}
     \vspace{-0.6cm}
    \caption{Steps to discover features in LLMs using sparse autoencoders (SAEs). The idea is to train a sparse autoencoder to project activations of an LLM to a sparse combination of features picked from a very large set of possible features (steps 1 and 2). Often---but by no means always---these features end up corresponding to human-interpretable concepts (which one can determine with step 3).}
    \label{fig:sae}
\end{figure}

To “disentangle” these features, MI researchers have developed tools like sparse autoencoders (SAEs).\footnote{Note that the reliability of sparse autoencoders is still debated in the literature. The central issue is whether SAEs actually discover a model’s features. In specific instances, SAEs exhibit limitations: features may fail to generalize beyond the training data \parencite{sharkey_open_2025}, become misaligned with actual features through “absorption” or “splitting” \parencite{chanin_is_2024, leask_sparse_2025}, suffer misinterpretation during manual inspection, or capture mere statistical artifacts \parencite{heap_sparse_2025}. However, solutions for these issues are already being proposed \parencite{makelov_towards_2024, bussmann_learning_2025, bayat_steering_2025}.} An SAE is an auxiliary neural network trained to decompress the dense, superimposed activation pattern of an LLM layer into a sparse, interpretable one where each active neuron corresponds to a single feature, rendering the activation pattern “monosemantic.” The SAE is trained to take the activation values of an LLM layer as input and to encode them in compressed form in a single hidden layer, but in such a way that the SAE can then still accurately decode its compressed representation to reproduce the original activation values (see Fig. \ref{fig:sae}).

In contrast to traditional autoencoders that have a small hidden layer to act as bottleneck and force compression, however, the SAE is given a hidden layer consisting of far more neurons than the LLM layer it must encode and decode. This provides a higher-dimensional latent space where features can, in principle, be embedded distinctly, rendering superposition unnecessary. But it also means that the SAE could simply copy the LLM’s activation values into its more capacious hidden layer. To prevent this, and to encourage the SAE to use a monosemantic encoding, a crucial constraint is imposed on the hidden layer: the way it encodes the LLM’s activation values must be \textit{sparse}, meaning that the SAE gets penalized unless most of its neuron activations are at or near zero. The idea is that if an LLM layer actually only relies on a small number of features when processing an input, the SAE, when forced to be sparse, will be driven to identify these most important features and represent each of them with a single, “monosemantic” neuron. 

Researchers can then inspect these monosemantic neurons—for instance, by finding text inputs that maximally activate them—to understand what human-interpretable concepts the LLM has learned \parencite{templeton_scaling_2024}. Thus, SAEs act like a computational “comb,” separating the entangled features in an LLM layer. It was using such an SAE that the “Golden Gate Bridge” feature and other features were discovered in state-of-the-art LLMs.

In summary, superposition is a critical mechanism that allows LLMs to make the most of their vast representational capacity, forming the basis for nuanced conceptual understanding. However, superposition also obscures the internal workings of these models. This makes techniques like SAEs vital tools in the MI toolkit, helping to overcome the challenges posed by superposition and elucidate the rich landscape of features that LLMs develop.

\subsection{The Transformer Architecture: Deepening Conceptual Understanding} \label{subsection:2.4}

Having established that LLMs develop a rich repertoire of features (2.1), represent them as directions in latent space (2.2), and can store a vast number of them via superposition (2.3), the crucial next question is how these features are \textit{used} to achieve relevant conceptual understanding in context. The transformer architecture plays a crucial role here by enabling the model to dynamically select and integrate information from different parts of an input sequence. Accordingly, we focus on decoder-only, causal masking, transformer-based LLMs---the standard architecture used notably in OpenAI’s GPT, Google’s Gemini, and Anthropic’s Claude models.\footnote{This is documented for GPT-2/3 \parencite{radford2019language, brown2020language} and Gemini models \parencite{geminiteam2025geminifamilyhighlycapable}, and implied for Claude models in Anthropic's research publications \citep[e.g., for Haiku 3.5][, see Fig. 1]{lindsey_biology_2025}. GPT-4/5 models have undisclosed architectures \parencite{openai2024gpt4technicalreport, openai2025gpt5systemcard}.}

To convey an intuitive picture of how transformer-based LLMs work internally, we adopt a \textit{single}-\textit{token}, \textit{sequential} \textit{perspective} on the transformer architecture: we zoom in on how an LLM handles just one token as it passes through the LLM’s many layers. Since the model applies the same set of operations to every token in a sequence, understanding what happens to one token gives us a good picture of how the entire model functions.

In tracking a token’s journey through the model, the central pathway to focus on is the “residual stream” associated with that token \parencite{elhage_mathematical_2021}---it constitutes the model’s continuously evolving understanding of a token from the initial input to the eventual output. This begins with a token entering the model and being converted into a numerical representation or vector---this is called \textit{embedding}. This initial embedding is a first shot at interpreting the token, containing general information about it without considering the words around it. It amounts to giving the model a basic and entirely \textit{context}-\textit{free} definition that is useful for simple predictions, such as guessing what might follow the token based on common two-word patterns (“apple” is frequently followed by “pie”). Additionally, positional embeddings are added at this stage to encode information about the token’s position within the sequence.

\begin{figure}[h]
    \centering
    \includegraphics[width=\linewidth]{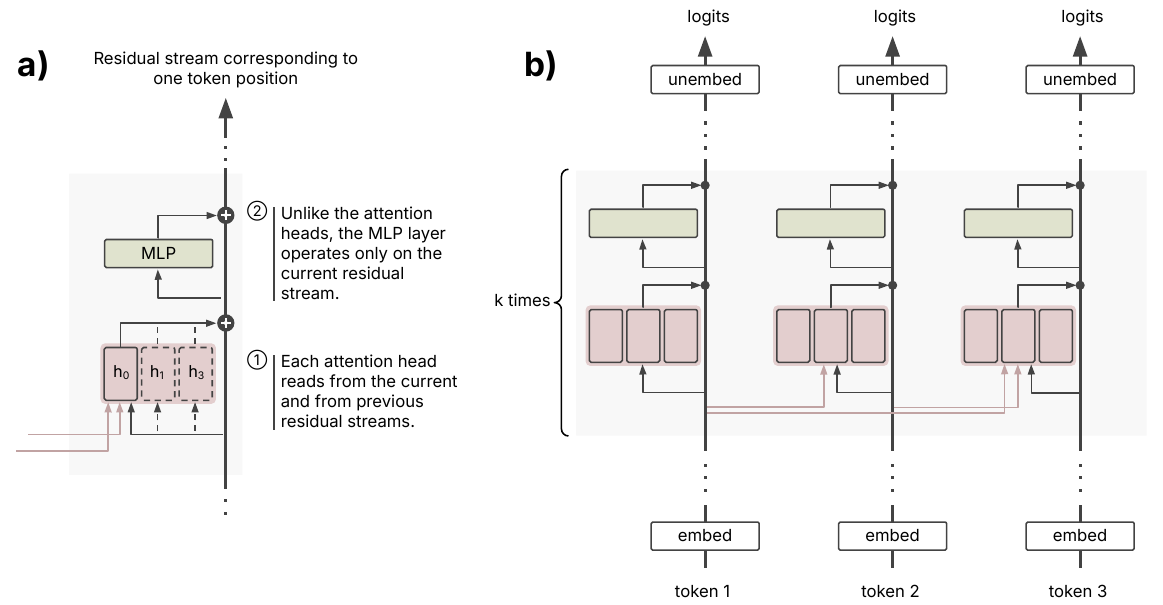}
     \vspace{-0.6cm}
    \caption{General overview of a decoder-only, transformer-based LLM. (a) Each transformer block consists of an attention layer, where attention heads operate in parallel, and an MLP layer. Both the attention heads and the MLP layer add to the embedding that will ultimately produce token predictions. The main information stream to which the transformer blocks contribute is called the residual stream. There is one for each token. (b) At each token, the model computes logits---representing likelihoods of the upcoming token---by passing through an embedding stage, $k$ transformer blocks, and an unembedding stage. }
    \label{fig:transformer}
\end{figure}

The resulting numerical representation of the token then passes through a certain number of processing units; in a transformer architecture, these units are called \textit{transformer} \textit{blocks}. Each transformer block consists of two successive layers (of which more below): an \textit{attention} layer and a \textit{multi}-\textit{layer} \textit{perceptron} (MLP) layer (see Fig. \ref{fig:transformer}a). Both kinds of layers can \textit{read} \textit{from} and \textit{write} \textit{to} the residual stream via their weight-defined linear projections. Before a token representation runs through an attention layer or an MLP layer, a so-called \textit{layer normalization} is applied. This is like adjusting an audio signal to a standard level, making sure it is neither too loud nor too quiet.

Once the representation of the token has passed through a certain number (\textit{k}) of transformer blocks, the embedding process is reversed at the \textit{unembedding} stage, which produces \textit{logits}---raw scores indicating the likelihood of each next token (see Fig. \ref{fig:transformer}b). These logits are then converted into a probability distribution summing to one via the softmax function before the model samples from likely next tokens to generate the next chunk of its response.

Attention layers play a crucial role in allowing the model to dynamically select and make effective use of the features it has learned. Each attention layer is composed of multiple “attention heads” operating in parallel. The fundamental job of an attention head is to retrieve relevant information from the residual streams of \textit{previous} tokens and integrate it into the residual stream embodying the model’s evolving understanding of the \textit{current} token (see Fig. \ref{fig:transformer}). This enables the model to transition from a \textit{context}-\textit{independent} to a \textit{context}-\textit{aware} understanding of the token: without context, “bank” might be taken to refer to a financial institution, but the nearby occurrence of “water,” “mud,” and “fish” indicates that the correct interpretation in this context is “riverbank.” Similarly, without context, the token “bridge” might activate a general “bridge feature.” However, if previous tokens include “Golden Gate” or “San Francisco,” attention heads can identify these cues, retrieve the more specific “Golden Gate Bridge” feature from those earlier token positions, and emphasize it in the current representation of “bridge.”

\begin{figure}[h!]
    \centering
    \includegraphics[width=\linewidth]{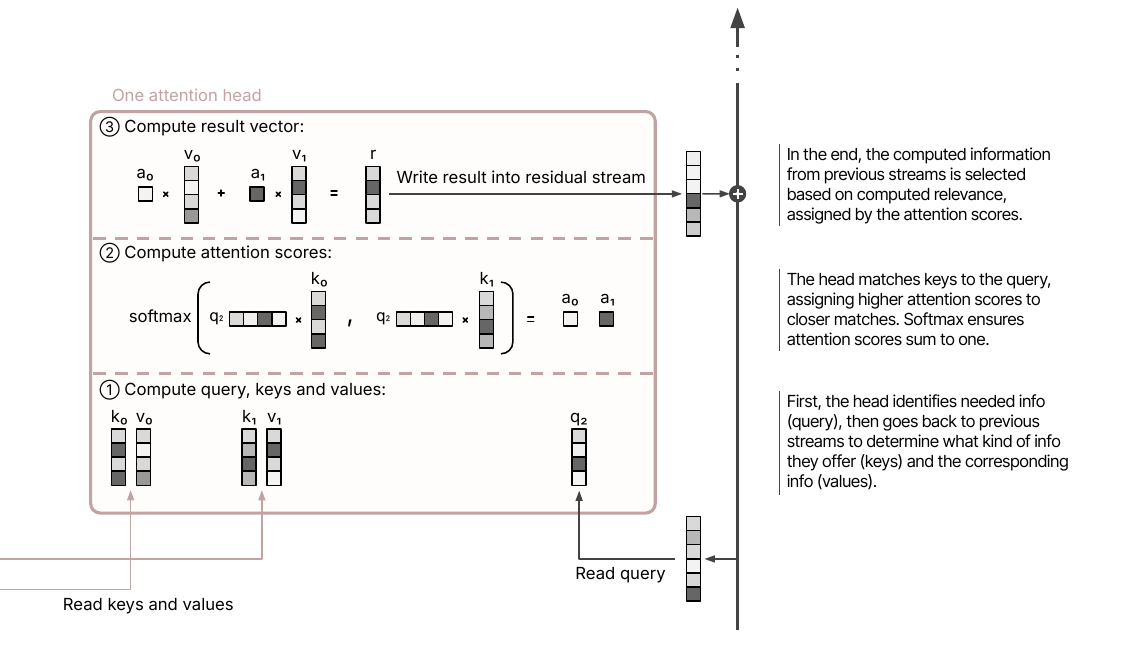}
     \vspace{-0.7cm}
    \caption{The operation of a single attention head. This allows the model to weigh the importance of different tokens in a sequence when processing a specific token. The process begins when the head generates a Query vector ($q_2$) for the current token (at position 2), representing the information it needs. Simultaneously, it generates Key vectors ($k_0, k_1$) and Value vectors ($v_0, v_1$) for previous tokens in the sequence. The key $k$ acts as a label for the information offered by a token, while the value $v$ contains the actual content. To determine relevance, the query $q_2$ is compared with each key ($k_0$ and $k_1$), producing raw similarity scores. These scores are then normalized by a softmax function to create the final attention scores ($a_0, a_1$), which are weights that sum to one. Finally, a result vector ($r$) is computed by taking a weighted sum of the value vectors ($v_0$ and $v_1$) using their corresponding attention scores. This result vector, which is a blend of relevant information from other tokens, is then added back into the residual stream.}
    \label{fig:attention}
\end{figure}

Each attention head learns to perform a specialized type of information retrieval from the residual streams of preceding tokens (see Fig. \ref{fig:attention}).\footnote{Because transformer models are designed to process information in parallel to optimize training, however, an attention head at a certain depth within the network---say, at layer 3---can only draw on the residual streams of preceding tokens at the same depth within the network, i.e. at layer 3. This imposes counterintuitive constraints on the information flow: an attention head may sometimes be forced to compute information for itself that previous residual streams also compute, but only at later layers, thereby depriving the attention head of access to the relevant information \citep[see][]{nostalgebraist_information_2024}.} The exchange of information does not happen at random, however. Through the attention mechanism, the head determines what information to seek and identifies which residual streams contain it. The attention head \textit{reads} from residual streams in three distinct ways. First, from the current stream, it forms a \textit{query} specifying the kind of information it is looking for. Second, it computes “keys” for all token positions in the context, which advertise the kind of information available at those positions. When a query from the current token strongly matches a key from a previous token, a high “attention score” is generated.\footnote{This attention score is obtained by first computing the dot product between the query and key vectors, followed by a softmax (see Fig. \ref{fig:attention}). These two operations introduce crucial non-linearities into the attention mechanism---enabling, as in the activation functions used in MLPs, a form of \textit{selectivity}.}  Third, this score then determines how much of the “value”---the actual information available to be passed along---from that previous token’s residual stream is copied and added to the current token’s stream.

Consider the sequence “A scared, winged creature.”\footnote{Our presentation of this example is inspired by Grant Sanderson’s video on LLMs, which we recommend as a complement to this section (\citeyear{sanderson2024llms}).} The residual stream for the “creature” token has to lead to a token prediction, presumably of a verb. One attention head might specialize in retrieving adjective information. Its query might implicitly ask: “What properties are being ascribed to the main subject?,” while its keys would advertise tokens that happen to be adjectives. These tokens would then get a strong match and the attention head would assign a strong attention score to them, allowing for their corresponding value vectors (which would simply be compressed versions of their residual streams) to be added to the “creature” residual stream, enriching it with these attributes and helping to predict the verb to follow (a \textit{scared}, \textit{winged} creature is likely to “fly,” “dart,” “huddle,” or “shriek,” but unlikely to “stroll,” “debate,” or “evaporate”).

This mechanism allows an LLM to figure out which features are most relevant for interpreting the current token in its particular context and to dynamically differentiate it from other contexts. If a text discusses the “sharpness of the debate” and later mentions “the bridge’s sharp lines,” for example, attention heads can differentiate these contexts: in the first context, heads might link “sharpness” to features for “intellectual discourse,” while in the second, they might link it to features for “physical form” or “visual properties.” Other heads might specialize in linking pronouns to their antecedents, or properties to the objects they describe, all by selectively attending to and moving feature activations across residual streams.\footnote{Thanks to \textit{computational} superposition, moreover, one attention head can perform multiple functions at once. Computational superposition differs subtly from the \textit{representational} superposition discussed above. While the latter is a property of the \textit{activations} within a network, computational superposition is a property of the \textit{parameters} within a network component (such as an attention head). Whereas representational superposition occurs when several features are represented in the same activations, computational superposition occurs when several distinct operations are simultaneously performed by the same network component. The Query-Key matrices of a single attention head, for example, do not just learn to attend to one type of relationship; rather, they are optimized to operate on input vectors that hold a large number of features in superposition.}

In essence, attention heads act as sophisticated routers and information integrators across residual streams. They scrutinize the features active across the context (as stored in the residual streams of previous tokens) and, guided by the current token’s characteristics, selectively draw upon these surrounding features. This context-sensitive retrieval and routing of information is essential for nuanced conceptual understanding.

While attention heads are crucial for gathering and routing contextually relevant information, the \textit{multi}-\textit{layer} \textit{perceptron} (MLP) layers, which follow the attention layers in each transformer block, play an equally vital role in deepening conceptual understanding. If attention heads are about identifying and carrying over relevant information \textit{across} residual streams, MLP layers perform computations \textit{within} the current token’s residual stream.\footnote{Note that as an attention head also performs in-place projections when it computes the value vector and when it writes back into the residual stream, it can also be seen as performing in-place combination and information-recall. However, MLP projections are larger and MLPs constitute more than two thirds of an LLM’s parameters. It thus makes sense to see them as the primary locus of such in-place computations.} They focus on \textit{processing}, \textit{refining}, and \textit{enriching} that information.

More precisely, MLPs can be seen as performing two functions. On the one hand, they combine the diverse pieces of information retrieved by attention layers. For instance, once an attention head has brought features for “Golden Gate” and “Bridge” into the current token’s stream, an MLP layer must combine these distinct features to activate a higher-level “Golden Gate Bridge” feature more strongly.

On the other hand, MLPs augment the resulting combination with associated knowledge recalled from the model’s learned parameters \parencite{nanda_fact_2023, geva_dissecting_2023, chughtai_summing_2024}. This information recall can take many forms. But the one that is most fundamental to conceptual understanding is the ability to connect a strongly activated feature to other features that help define and identify it. Thus, if the “cat” feature is strongly activated within the current residual stream, an MLP layer might recall associated information stored within its weights, enabling the LLM to connect the “cat” feature to other features that are definitive of being a cat, such as “mammal,” “pointy-earedness,” and “furriness” (we explain how such feature-activated information recall works in section \ref{subsection:3.1}). 

This recall process significantly enriches the model’s features, making them more nuanced and useful for predicting subsequent tokens that depend on the ability to exploit these conceptual connections. This is fundamental to an LLM’s ability to move beyond superficial token-co-occurrence patterns and leverage conceptual connections among features. And grasping these conceptual connections amounts to a form of understanding. “To have understanding,” Jonathan Kvanvig writes, is notably “to grasp … conceptual connections” (\citeyear{kvanvig2018knowledge}, p. 699). There is a rich philosophical tradition holding that to understand a concept is, above all, to master its connections to other concepts that obtain in virtue of the content of the concept---one does not grasp the concept “bachelor” unless one grasps its connections to the concepts “man” and “unmarried.”\footnote{Such conceptual connections figure particularly prominently in the inferentialist tradition (Sellars, Dummett, Harman, Brandom, Block, and others, which has been thought to be especially well-suited to making sense of LLMs \parencite{piantadosi2022meaning, piantadosi2024concepts}. But conceptual connections are also key to the much older idea of “grammatical truths” (Wittgenstein), “analytic truths” (Kant), or “truths of reason” (Leibniz).} By empowering LLMs to recall analogous connections between features, MLPs significantly strengthen the case for the attribution of something like conceptual understanding to LLMs. Moreover, both the combination of features and the recall of associated information often occur in superposition within the MLP layers, further leveraging the model’s dense representational capacity.

Importantly, passing through a transformer block only \textit{adds} to the information in the residual stream, without erasing or distorting information from earlier layers. As an information package enters a transformer block, it is processed by the attention layer and the MLP layer, but a copy is also sent directly to the end of the transformer block (via a so-called “skip connection”) and added to the processed information. The block thereby learns to compute the \textit{difference} or “residual” between the original input to the block and its desired output, but without destroying the original input.

This contrasts with classical multi-layer perceptrons without skip connections, where activations are thoroughly and irretrievably transformed from one layer to the next. This might lead to a loss of important original information, which makes it harder for the network to learn (this is related to the “vanishing gradient problem”). Preserving the original information that enters a transformer block avoids this problem, giving the network the freedom to pass information through a transformer block \textit{unchanged} where necessary. This gives the model the freedom to be selective about which layers it processes information through instead of having every token be affected by every layer. And in fact, it has been observed that MLPs and attention layers are highly selective, each interacting with a narrow, dedicated subspace of the residual stream---introducing a kind of division of labor \parencite{elhage_mathematical_2021, su2025conceptscomponentsconceptagnosticattention}. The model can pick up on more complex patterns as a result.

The realization that passing through a transformer block only adds to the information in the residual stream has a convenient upshot: we need not consider a transformer model as a \textit{series} of latent spaces (as in Fig. \ref{fig:latent_space}), but can see the different transformer blocks as successively writing into a \textit{single}, \textit{fixed} \textit{latent} \textit{space} in which the residual stream is progressively updated.

The assumption of a single, fixed latent space is supported by an MI technique called “logit lens” \parencite{nostalgebraist_interpreting_2024}, which works by treating an activation vector in an intermediate layer as if it were the final one and running it through the unembedding matrix to see what next-token predictions would be made based on the information accumulated up to that point. By applying the logit lens at each layer, researchers can create snapshots of the model’s evolving predictions. They found that the model did not produce radically different predictions at each layer, but successively refined them, supporting the view of the residual stream as a cumulative computational workspace.

In summary, the transformer architecture provides the mechanisms necessary for LLMs to dynamically utilize and refine their learned features. Attention layers act as dynamic routers, scrutinizing the context and selectively drawing relevant information from across the input sequence into the current token’s representation. MLP layers process this contextualized information, combining lower-level features to activate higher-level features and enriching the resulting representation by recalling associated information. This iterative process of contextual selection by attention and conceptual refinement by MLPs, repeated through multiple transformer blocks, allows LLMs to build up an increasingly nuanced and coherent representation of the input sequence. It is this dynamic interplay that enables them to effectively keep track of entities and properties, resolve ambiguities, and ultimately manifest a robust form of conceptual understanding by connecting diverse manifestations of something in a contextually appropriate manner.

\section{State-of-the-World Understanding} \label{section:3}

The previous section argued that LLMs connect diverse manifestations of things by unifying them under a feature, much as we do by subsuming things under concepts. Accordingly, we called this foundational capacity “conceptual understanding.” We saw that conceptual understanding already involves connecting a feature (e.g., “cat”) to other features (e.g., “mammal,” “pointy-earedness,” and “furriness”). However, a more sophisticated level of understanding emerges once a model goes beyond these definitive connections (the \textit{a} \textit{priori} connections to other features that obtain in virtue of a feature's content) and starts to represent how different features are empirically connected in the world (its \textit{a} \textit{posteriori} connections to other features that obtain in virtue of how the world contingently is).\footnote{We leave aside here the longstanding philosophical debate over whether that distinction can be drawn sharply in the case of concepts. Even if the distinction is not clear-cut, both sides agree that there is a significant difference between these two ways of connecting concepts to each other.} This is what we term \textit{state}-\textit{of}-\textit{the}-\textit{world} \textit{understanding}.

State-of-the-world understanding involves the LLM forming an internal model of how distinct features relate to each other to constitute facts. This allows the model to learn factual associations between features and output sentences like “Marie Curie was a physicist,” not merely as statistically likely sequences of tokens, but as outward expressions of an internal organization that links these features. This ability to learn factual associations between features aligns with a central tenet in the philosophical literature on understanding: that understanding is factive, as it relies on connections that obtain in the actual world \parencite{baumberger2017understanding}.

There are different kinds of state-of-the-world understanding. In what follows, we distinguish between \textit{static} state-of-the-world understanding---the learning of unchanging factual associations between features (“Marie Curie was a physicist”)---and \textit{dynamic} state-of-the-world understanding, which involves updating an internal model of the world in response to changes in the world. We first provide an intuitive explanation of how LLMs might achieve static state-of-the-world understanding (\ref{subsection:3.1}) before turning to an example suggesting that they are also capable of dynamic state-of-the-world understanding (\ref{subsection:3.2}).

\subsection{Learning Factual Connections} \label{subsection:3.1}

To learn facts is to learn how different entities and properties relate to each other in the real world. Having concepts is one thing, but understanding that “Marie Curie was a physicist” requires connecting those concepts in a way that reflects reality. How does an LLM learn to draw these factual connections?

The answer lies primarily in the MLP layers of the transformer. While we have seen how MLPs help refine representations by recalling conceptual associations, their other crucial role is to enable the recall of factual associations. An MLP layer can be thought of as a sophisticated switchboard that has learned to connect thousands of different features. The specific wiring of this switchboard—encoded in the model’s weights during training—represents the factual connections between features that the model has learned.

\begin{figure}[h!]
    \centering
    \includegraphics[width=\linewidth]{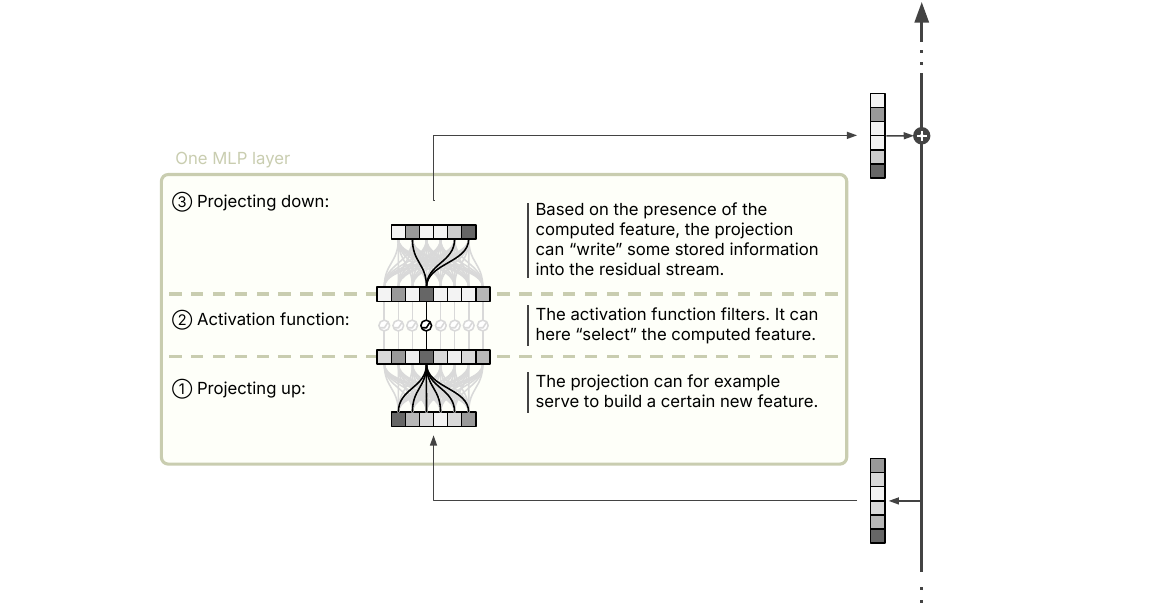}
     \vspace{-0.7cm}
    \caption{The MLP Layer as a feature-combiner and feature-activated information recaller. The MLP performs three key operations: (1) it projects a vector from the residual stream into a higher-dimensional space; (2) it applies a non-linear activation function (represented here by an encircled ReLU function) to filter the projected activations; and (3) it projects the filtered activations back to the original dimensionality to be added to the residual stream. The darker lines in the diagram illustrate a hypothetical idealized example where the up-projection combines input activations to compute one specific feature (such as “Golden Gate Bridge”), the activation function acts as a gate, determining if the feature’s activation is strong enough to be passed on, and the down-projection recalls associated information (such as “is in San Francisco”), to be added back to the residual stream via a skip connection (shown by the $\oplus$ symbol).}
    \label{fig:mlp}
\end{figure}

LLMs are capable of encoding factual connections between features via the linear projections of their MLP layers. These layers typically consist of two linear projections: an “up-projection” that expands the dimensionality of the representation, followed by a non-linear activation function that acts as a filter, and then a “down-projection” that restores the representation to the original size of the residual stream (see Fig. \ref{fig:mlp}). The fixed linear weights within these projections directly connect certain features to others, thereby enabling the encoding of factual associations. 

To understand how this works, consider an idealized case where features align perfectly with individual neuron directions. Imagine, for example, that an MLP’s first linear projection successfully combines features for “Golden Gate” and “Bridge” into a single, neuron-aligned “Golden Gate Bridge” feature. That feature then passes through a non-linear activation function, which effectively filters out accidental or spurious activations and only allows sufficiently strong activations of the “Golden Gate Bridge” feature to propagate forward.

Now, the second linear projection can recall associated facts. If the output layer of the MLP possesses a neuron-aligned “in San Francisco” feature, a single learned weight connecting these two neurons—if set to a high value—can directly encode the factual connection that “the Golden Gate Bridge is in San Francisco.” When the “Golden Gate Bridge” feature activates, this weight ensures that the “in San Francisco” feature is strongly activated in the output as well, thereby effectively enabling the LLM to “recall” a fact.

In practice, features rarely align with single neuron-directions. Yet the same principle applies: the MLP layer learns a mathematical transformation (a linear projection) that takes the vector representing the “Golden Gate Bridge” direction and reorients it so that the output vector has a strong component along the “in San Francisco” direction. It can do this for multiple facts at once. Looking at Figure \ref{fig:mlp}, for example, the strong activation of the Golden Gate Bridge feature (node 4 before the down-projection) might activate three facts—“in San Francisco,” “opened in 1937” and “designed by Joseph Strauss”—in the next layer (nodes 2, 5 and 6). The weight matrix of the MLP would then learn to transform the “Golden Gate Bridge” direction into an output vector that can be decomposed into three feature directions—perhaps $0.9 \times$ “San Francisco” direction $+ 1.0 \times$ “completed in 1937” direction $+ 0.8 \times$ “designed by Joseph Strauss” direction. 

This highlights the capacity of LLMs to recall associated information via factual as well as conceptual associations. Traditionally, deep learning models have been seen as cascades of feature combinations, successively assembling lower-level features into higher-level ones. The Convolutional Neural Networks used for image recognition provide the paradigmatic example of this---first detecting lines, edges and simple patterns, then assembling them into more complex patterns, and eventually arriving at objects. In LLMs, MLPs do act as such feature-combiners (e.g. when combining the “Michael” and “Jordan” features); but they also go significantly beyond this, serving as \textit{feature}-\textit{activated} \textit{information} \textit{recallers} that respond to the strong activation of a feature by augmenting and enriching it with the conceptual and factual information stored in their parameters \parencite{BeckmannManuscript-BECDLM-2}.

This mechanism of factual recall through MLP layers has been empirically validated by MI research. Studies using targeted interventions confirm that MLP layers are the primary locus of information enrichment \parencite{geva_dissecting_2023, nanda_fact_2023}. There is also strong evidence suggesting that information typically gets recalled in the residual stream of the subject—in our example, the “Golden Gate Bridge” stream—thereby making it available for upcoming residual streams that might need it \parencite{geva_dissecting_2023, nanda_fact_2023, ameisen_circuit_2025, lindsey_biology_2025}. Furthermore, as we mentioned in footnote 17, attention heads also recall information during their routing task, via their direct linear projections: the value projection and the final output projection \parencite{geva_dissecting_2023,chughtai_summing_2024}.

By combining factual recall with other simple mechanisms, LLMs can also deal with more complex cases of predication. For example, MLPs can deal with negation by decreasing feature activations. If the input mentions the Golden Gate Bridge, an MLP might reduce the activation of the “Los Angeles” feature---helping the model encode that “the bridge is not in Los Angeles” (see §3.7 in Ameisen et al., \citeyear{ameisen_circuit_2025}).

MLPs might also handle predication involving temporal properties using conditional gating of recall: the model can learn to retrieve a fact \textit{only when certain conditions are met}. For example, the activation of a “year 1900” feature could inhibit the usual association of “Golden Gate Bridge” with “San Francisco,” allowing the model to express that the bridge was not in San Francisco \textit{in 1900}. To our knowledge, this specific temporal-gating hypothesis has yet to be tested, but there is evidence that LLMs encode time-related features \parencite{gurnee_language_2024} and often implement more complex cases of predication via feature-level inhibition (e.g., Lindsey et al., \citeyear{lindsey_biology_2025}, §8)

Through the learned weights of its MLP layers and attention heads, the transformer architecture thus provides a clear mechanistic basis for a form of state-of-the-world understanding. It enables LLMs to move beyond just encoding features to encoding the factual associations between them.

\subsection{Internal Models of External States: The Othello-GPT Findings} \label{subsection:3.2}

The mechanisms discussed thus far concerned static or unchanging facts. But the world is not static; it continually changes. Can an LLM dynamically update its internal representations of evolving states of affairs?

Research on Othello-GPT provides a compelling affirmative example. Othello-GPT is a language model based on GPT-2 that was trained to generate legal moves in the game of Othello by predicting the next move. The model was trained solely on game transcriptions, without ever seeing the board or even the game’s rules. Despite this limited input, when provided with a sequence of moves from a game, Othello-GPT reliably outputs only moves that are legal for the current board position. This ability generalizes to unseen sequences. The model was tested on game sequences that had been systematically withheld from the training data (e.g., all the sequences downstream of the opening move C5, which represents a quarter of the game tree). Even when traversing these unfamiliar branches of the game tree, the model successfully predicted legal moves, demonstrating that it was not simply relying on memorized game sequences \parencite{li_emergent_2023}.

To appreciate the cognitive demands this places on the model, it is necessary to understand just how dynamic board configurations in Othello are. The game is played on an 8x8 grid where two players, black and white, alternate placing discs. A move is legal only if it horizontally, vertically, or diagonally sandwiches one or more of the opponent’s discs between the newly placed disc and an existing disc of the player’s own color (Fig. \ref{fig:othello_gpt}a). All sandwiched discs are immediately flipped to become the current player’s color. This makes the identity of the pieces on the board highly fluid: unlike in chess, where the fact that there is currently a black knight on D4 can straightforwardly be read off the move history, in Othello, a disc placed on E5 three moves ago is no guarantee that a black disc still occupies that square. A single move can completely reconfigure the board by flipping multiple discs along multiple directions at once. 

Given Othello-GPT’s success at generating moves that are legal in the current position, the question is whether the model achieves this by exploiting superficial statistics or by continuously maintaining an internal representation of the board state. When Li et al. (\citeyear{li_emergent_2023}) studied this, they found striking evidence that Othello-GPT indeed maintains an internal representation of the external state of the board. A vivid way into these findings is an analogy made by Li (\citeyear{li_emergent_2023}): suppose you regularly play Othello with a friend, calling out each move as you make it (“E3,” “D3,” “C5,” etc.). After many games, a crow outside the window, which could hear everything, but not see the board, starts calling out moves. To your amazement, these turn out to nearly always be legal next moves given the current state of the board. You naturally assume that the crow is just reproducing the sounds it was exposed to, perhaps guided by a superficial sense that certain moves (typical opening moves, for instance) tend to come earlier while others tend to come later in the sequence. Yet after a few more games throughout which the crow calls out only legal moves, you are sufficiently intrigued to inspect the crow’s perch. You find a grid-like arrangement of two sorts of seeds showing the exact state of your ongoing Othello game. Coincidence? You chase away the crow and rearrange some of the seeds. When the crow returns, it calls out a move that is legal only in the rearranged position. At this point, it becomes difficult to deny that the crow is doing more than mimicry based on superficial statistics. It must be keeping track of the board state and basing its calls on that.

To determine whether Othello-GPT was actually maintaining an internal representation of the current state of the board---an “emergent world model,” as Li et al. (\citeyear{li_emergent_2023}) call it---the researchers employed a technique known as \textit{probing}. Probing, established as a key method for mechanistic interpretability by Alain and Bengio (\citeyear{alain_understanding_2017}), serves to detect if and where a given feature is encoded in a model.\footnote{For a philosophical discussion of when linear probes reveal LLMs’ beliefs, see Herrmann and Levinstein (\citeyear{levinstein2024}).} One starts by selecting a target feature and a layer to investigate. One then trains a separate model (the “probe”) to predict, based on the chosen layer’s activation pattern when presented with an input, whether the feature is present in that input. The probe learns to do this by seeing lots of examples of both activation patterns produced by inputs explicitly labeled as \textit{containing} the feature and activation patterns produced by inputs explicitly labeled as \textit{lacking} the feature. As readers of Wittgenstein (\citeyear{Wittgenstein1953-WITPI-4}, §§28–33) on the indeterminacy of ostensive definitions will know, there is a challenge involved in selecting positive and negative examples that unambiguously exemplify the target feature as opposed to its many correlates: a collection of inputs mentioning politicians instantiate the feature “politician,” but the contrast class of negative examples needs to be carefully chosen to ensure that the probe does not learn to probe for the “political” feature, or the “person” feature, which are equally instantiated by the positive examples \citep[see][]{gurnee_finding_2023}. Once suitable training data have been assembled and labeled, however, one can pass the inputs through the model to be investigated, gather the activation patterns at the chosen layer, and see if the probe can learn to predict the feature’s presence from these activation patterns alone. If the probe can learn to do this accurately on new, unseen inputs, this provides strong evidence that the feature is encoded in that layer.

\citeauthor{li_emergent_2023} trained probes to predict the state of each of the 64 board squares—whether it was empty, black, or white—solely from the model’s internal activations. Their initial attempts using linear probes yielded high error rates. This ran counter to the Linear Representation Hypothesis (LRH), suggesting that the model did not store the board state in a linear, easily intelligible format. However, when they switched to non-linear probes (specifically, 2-layer multilayer perceptrons), accuracy soared, with error rates dropping as low as $1.7\%$. This indicated that a representation of the board was indeed encoded within the network, albeit in a complex form.

Yet the initial failure to find simple linear encodings of the board state left the door open to a more skeptical interpretation: if a model’s internal states are highly entangled and non-linear, a skeptic might argue that it has not really learned to represent the board at all. Consider an analogy with jet streams in a climate model. Even if no explicitly represented variables stand in for jet streams, they are still present in an emergent way in the dynamics---recoverable, in principle, by a classifier informed by the relevant background knowledge. Yet one would not want to say that the climate model “understands” jet streams. In the same way, a deep neural network may lack any explicit representation of the board state, even if that state can be decoded from its distributed activation patterns. The skeptic's point is that if one would not say that the climate model “understands” jet streams, one should not say that LLMs “understand” the board state in Othello either.\footnote{We are indebted to an anonymous reviewer for this way of phrasing the skeptical challenge.}

This skeptical challenge became increasingly unconvincing in light of subsequent analyses, however. A study by Nanda et al. (\citeyear{nanda_othello_2023}) provided the first counter-argument, revealing that the model’s internal representation of the board was simpler and more elegant than first thought. The trick was to reframe what to probe for. Instead of looking for features describing the board in absolute terms (“Black” vs. “White” discs), Nanda et al. hypothesized that the model might track pieces in relation to a player’s perspective (“Mine” vs. “Yours”). This time, linear probes were successful. To appreciate the import of this result for the skeptical challenge, one needs to understand how linear probes work.

Unlike the non-linear probes used by \citeauthor{li_emergent_2023}, linear probes operate under the assumption that the LRH holds true: whatever features the model learns are represented as lines or directions in its latent space. To visualize this, imagine the entire activation pattern that an input produces at one layer not as a list of neuron activations or numbers, but as a single point: the point picked out by the “activation vector” that this list of numbers defines. If a model has learned to treat a direction as corresponding to a feature, it will organize the inputs so that inputs \textit{with} that feature register as a point in one region of latent space and inputs \textit{without} that feature register as a point in a clearly distinct region.

\begin{figure}[h!]
    \centering
    \begin{minipage}{0.5\linewidth}
        \centering
        \includegraphics[width=\linewidth]{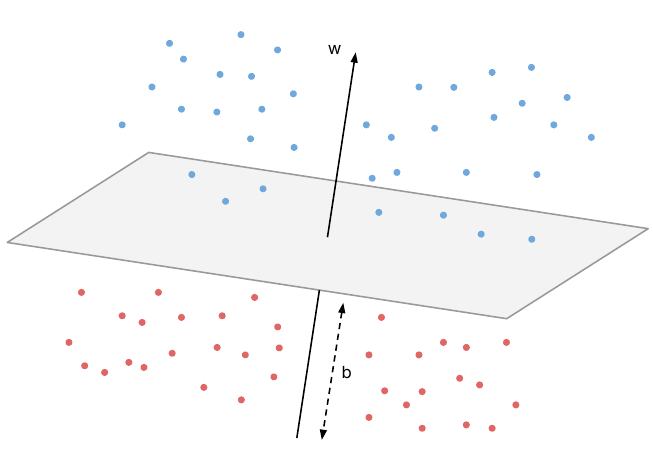}
    \end{minipage}%
    \hspace{0.5cm}
    \begin{minipage}{0.4\linewidth}
        \caption{Illustration of a linear probe’s training objective. The probe is trained on labeled data to find a hyperplane (defined by the weight vector $w$ and the bias term $b$) that separates the points generated by inputs in which the sought-after feature is present from the points generated by inputs in which the feature is absent.}
        \label{fig:linear_probe}
    \end{minipage}
\end{figure}

\begin{figure}[h!]
    \centering
    \includegraphics[width=\linewidth]{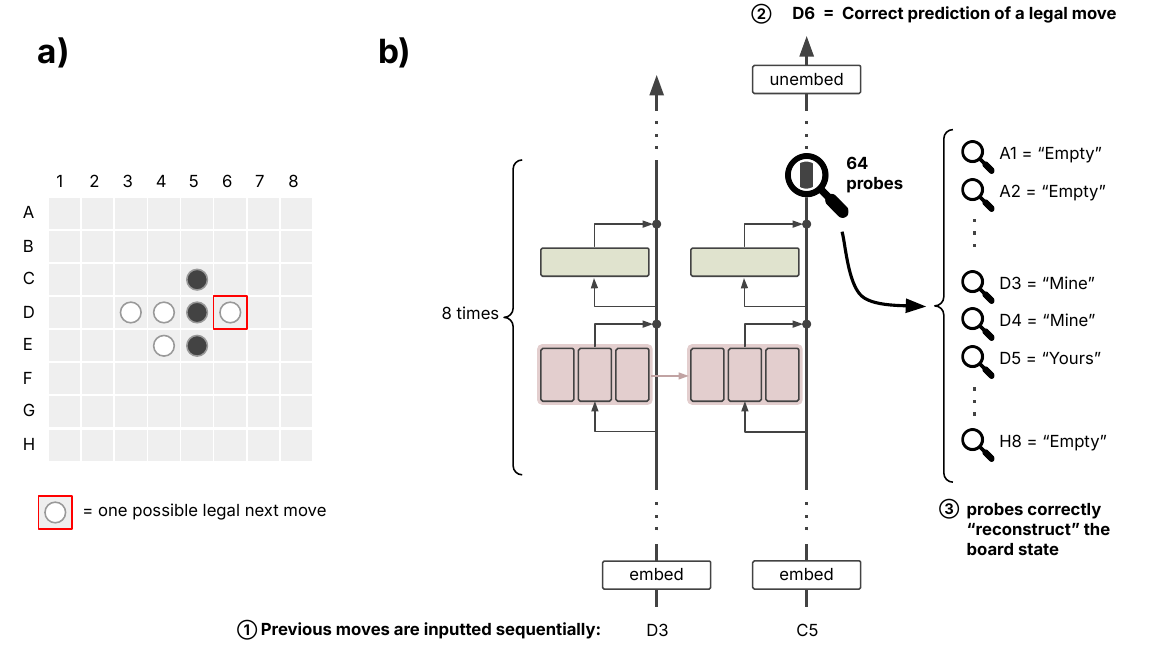}
     \vspace{-0.6cm}
    \caption{Probing board states from Othello-GPT. The model is asked to predict legal moves given the sequence of moves that led to the board state shown in (a). Legal moves have to use a disc of the player whose turn it is (here: black) and place that disc on a previously empty square in such a way as to “sandwich” the opponent’s discs (here: white) between the player’s own discs. This can be done horizontally, vertically, or diagonally—“D6,” for example, is a legal move for black here. (b) As the model processes previous moves, 64 separately trained linear probes each correctly predict the state of one of the 64 board squares from the residual stream (and this works at various locations of the stream, e.g., after 4, 5 or 6 transformer blocks).}
    \label{fig:othello_gpt}
\end{figure}

A linear probe seeks a flat dividing surface (in a high-dimensional space, this is called a \emph{hyperplane}) that cleanly separates the points generated by inputs with the feature from those generated by inputs without the feature (see Fig. \ref{fig:linear_probe}).\footnote{Alternatively, to stick with the image of vectors, one might say that the probe seeks a hyperplane that intersects with all and only the activation vectors that express the feature to a sufficient degree.} The probe learns how to orient the hyperplane by learning a weight vector $\mathbf{w}$ that points orthogonally away from the hyperplane. The direction of this weight vector corresponds to the feature that the probe is looking for. The probe also learns a bias term $b$, which positions the hyperplane by shifting it along the direction of the weight vector in relation to the origin. To determine whether a given activation vector $\mathbf{x}$ was produced by an input containing the feature, the probe measures how closely the activation vector aligns with the feature direction represented by the weight vector $\mathbf{w}$. Mathematically, this corresponds to calculating the dot product of the two vectors before adding the bias. The resulting score, determined by the formula $\mathbf{w} \cdot \mathbf{x} + b$, indicates which side of the hyperplane the point lies on: a positive score signifies that the feature is present, whereas a negative score indicates that it is absent.

The success of a linear probe shows that although a neural network uses non-linear computations, it has organized information such that features are arranged in a \textit{linearly separable way} (i.e. such that they can be cleanly separated by a hyperplane, the high-dimensional equivalent of a line). In other words, the network learns to place activations corresponding to “feature present” on one side of a hyperplane and those corresponding to “feature absent” on the other. A linear probe succeeds precisely because it is able to find this hyperplane.

Nanda et al. (\citeyear{nanda_othello_2023}) trained 64 distinct linear probes, one for each square of the Othello board. Each probe was designed to predict the state of its assigned square by examining the model’s activations in the last token’s residual stream, at the final step before the next legal move prediction (see Fig. \ref{fig:othello_gpt}b). Since each square has three possible states—it can be occupied by a black disc, a white disc, or remain empty—they trained each probe to learn three distinct weight vectors within the activation space, one for each of the possible square states (in this sense, each probe can be viewed as a combination of three distinct probes, as a single probe usually targets just one feature). In contrast to earlier studies that sought and failed to find linear encodings of “Black,” “White,” and “Empty” features, however, the key idea of Nanda et al. was to look for features that are relative to the perspective of the current turn’s player: “Mine,” “Yours,” and “Empty.” These features correspond to three directions in the residual stream’s latent space—the “Mine,” “Yours,” and “Empty” directions. To predict a square’s status, the probe projected the activation vector onto these three directions and identified the one yielding the highest score, thereby determining the most likely state.\footnote{A softmax function ensures the three projections sum to 1, converting them into probabilities.}

The results were striking. The probes could decode the state of every square on the board from the model’s activation vectors with near-perfect accuracy, suggesting that Othello-GPT organizes information in cleaner, more intelligible ways than initially feared. 

This is philosophically significant: a deep neural network would be capable of burying information in highly entangled, non-linear manifolds that defy simple linear probes. The fact that the full board state at any stage of the game can be recovered linearly suggests that the model has not only somehow learned to reconstruct the board state from the sequences of moves, but that it has learned to represent the board in a format that is intelligible to us, because it aligns with the logical structure of the game as seen from the perspective of a human player. Where the skeptic expects uninterpretable “spaghetti code,” we find an intelligible structure. Call this response to the skeptical challenge the \textit{argument from linearity}.	

Furthermore, preliminary evidence suggests that this internal structure itself mirrors the spatial structure of the board. This gives rise to a second response to the skeptical challenge that might be called the \textit{argument from isomorphism}.\footnote{See also Williams (\citeyear{WilliamsForthcoming-WILCSC-4}) for a complementary discussion of when such a structural correspondence can ground genuine representation.}

If the board states were encoded using conventional programming, we would not expect an isomorphism between the geometry of the board and the encodings inside the model. Board states would typically be stored in arbitrary memory locations—square (0,0) might sit at address 1000 while its neighbor (0,1) might sit at address 5000, with no meaningful spatial relationship between these addresses. Spatial relationships would be handled through explicit computational rules: to find neighbors of square (r, c), the program would check coordinates like (r-1, c), (r+1, c), and so on. The spatial structure of the board would be reflected only in these rules, not in how the information was organized inside the model.

Othello-GPT \textit{could} have done something similar: it could have scattered the directions it uses to represent board states at random and relied on additional computational machinery to track spatial relationships.

\begin{figure}[h!]
    \centering
    \includegraphics[width=\linewidth]{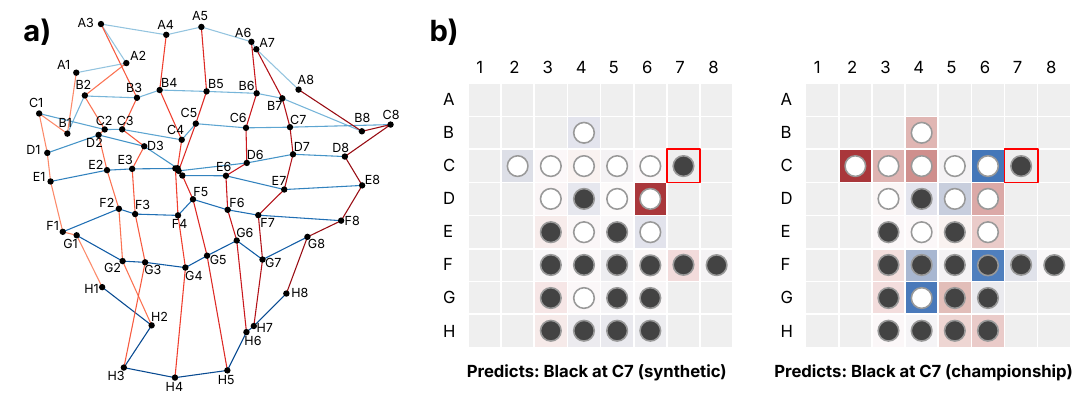}
     \vspace{-0.6cm}
    \caption{Empirical findings regarding the internal model of Othello-GPT. (a) A principal component analysis of linear probe directions reveals a spatial organization mirroring the Othello board structure. Rows are in blue (lightest blue = row 1) and columns in red (lightest red = column 1). PCA performed on all directions taken together at layer 7. (b) Latent saliency maps: the red dot marks the model’s top prediction; square colors indicate each square state’s contribution (red = high, blue = low; normalized). Contribution is measured by how much changing the internal representation of a square state lowers the prediction’s likelihood. Left: Othello-GPT trained on synthetic data. Right: Othello-GPT trained on championship data. The left map assigns high saliency only to the square required for legality: C7 is legal because it flips the white disc on D6. By contrast, the right map also highlights squares that do not affect legality directly and are far from the move, such as C2, suggesting sensitivity to broader board structure.}
    \label{fig:othello_results}
\end{figure}

Yet when Li et al. applied principal component analysis (PCA) to the directions identified by their probes—projecting them into a 3D space to reveal the three dimensions capturing the most variance—they found a spatial arrangement resembling the actual Othello board layout (\citeyear{li_emergent_2023}, Appendix). We replicated this analysis using \citeauthor{nanda_othello_2023}’s linear probes and two dimensions, and found an even more pronounced correspondence: the directions picked out by the probes to represent spatially adjacent squares are themselves adjacent in the projection, with rows forming roughly linear arrangements and columns doing likewise, reproducing the grid structure of the board itself (see Fig. \ref{fig:othello_results}a).\footnote{Some caveats apply. First, we do not carry out the last step of the typical MI pipeline: ensuring the identified signal is causally relied on by the model. Doing so would address a methodological concern raised by Li et al. (\citeyear{li_emergent_2023}), who observed some board structure—though considerably weaker and more jumbled—even when applying PCA to probes trained on random linear projections of Othello moves, suggesting spatial signal might partly arise from the probing method itself. Second, the model might encode and exploit spatial structure in many different ways internally: Yuan and Søgaard (\citeyear{yuan_revisiting_2025}), for example, found that token embeddings for Othello moves (e.g., “D4”) also cluster by spatial proximity. What we can conclude with reasonable certainty is that structural isomorphism is present in some form within the model's representations; demonstrating precisely how it is causally exploited awaits further investigation.}

Third, there is the \textit{argument from causal efficacy}: the board representation clearly plays a causal role in the move predictions. In line with philosophical accounts of representation in LLMs \parencite{yetman2025representationlargelanguagemodels, WilliamsForthcoming-WILCSC-4}, both teams of researchers went beyond observing correlations and demonstrated causal role via interventions. By manually “flipping” the representation of a disc from “Black” to “White” \parencite{li_emergent_2023} or erasing a previously played disc \parencite{nanda_othello_2023}, they were able to get the model to output moves that were legal only in the counterfactual board states and not in the original board states. This satisfies a robust definition of causal role: changing the internal representation systematically and predictably alters the model’s behavior. This shows that the model is not just predicting the next token based on previous tokens, but relying on its internal representation of the board.

The Othello-GPT findings thus provide a clear instance of dynamic state-of-the-world understanding in an LLM. Even in an LLM as simple as GPT-2, a sophisticated form of relational state-of-the-world understanding can emerge from next-token prediction. Further supporting this, Yuan and Søgaard (\citeyear{yuan_revisiting_2025}) extended these findings to more recent, larger LLMs, showing that despite architectural differences, these models converged on strikingly similar internal board representations.

This form of state-of-the-world understanding is \textit{dynamic} because Othello-GPT adjusts its emergent world model at inference---a feat significantly surpassing the Michael Jordan example, where the connection between features is set once and for all during training. Othello-GPT shows that LLMs can not only create a register of facts during training, but also update this register of facts at inference.

And crucially, Othello-GPT’s state-of-the-world understanding is relational and integrated. It does not treat each square state as an isolated fact, but constructs a representation of the board that enables it to attend to these facts \textit{together}, as a \textit{system} of facts. Evidence of this emerged when Li et al. (\citeyear{li_emergent_2023}) used “latent saliency maps” to visualize which squares played the largest role in informing the model’s top next-move prediction. Each square's contribution was measured by how much changing the internal representation of that square state lowered the prediction’s likelihood. They did this with two versions of Othello-GPT: one trained only on computer-generated sequences of random moves (synthetic data), the other trained on expert human games (championship data). While the version trained on synthetic data showed high saliency only for the squares that were strictly necessary to rendering a move legal, the version trained on championship data exhibited a much broader saliency pattern, mirroring how the moves of expert players depend on global features of the board (see Fig. \ref{fig:othello_results}b). Such a grasp of how elements of a system depend on each other is considered a hallmark of understanding \parencite{riggs2003balancing, grimm2011understanding, sep-understanding}. 

Nevertheless, we must also temper these conclusions with a recognition of the model’s limitations. Othello-GPT remains something of a “fruit fly” experiment: it involves a closed system with a finite state space (64 squares, 3 states per square) and perfect information. That an LLM can dynamically track the state of an Othello board does not guarantee that it could learn to track the state of the wider world, which is open-ended, noisy, and non-deterministic. We must be wary of over-extrapolating from these findings.

Moreover, the fidelity of this emergent state-of-the-world understanding is not uniform. \citeauthor{nanda_othello_2023} observed “end-game degradation,” where the model’s reliance on its internal representation of the board state seemed to falter in the final moves of a game. They hypothesized that as the board fills up and options become fewer, it becomes computationally cheaper for the model to switch from a complex world-modeling circuit to simpler heuristics. This suggests that LLMs may not be consistent state-of-the-world trackers. They may pragmatically toggle between modeling the state of the world and relying on shallow heuristics, depending on which circuit minimizes loss most efficiently. Thus, while the Othello-GPT findings suggest that LLMs \textit{can} maintain an internal representation of the state of the external world, they also warn us that they will not necessarily do so when a shortcut is available.

\section{Principled Understanding} \label{section:4}

While conceptual understanding allows a model to group diverse phenomena under a single feature, and state-of-the-world understanding enables it to learn contingent factual connections between these features, the highest level of our proposed hierarchy involves grasping the abstract principles or rules that unify a diverse array of facts. This is what we call \textit{principled} \textit{understanding}. An agent might know numerous individual facts yet fail to achieve principled understanding if they do not grasp what unifies them---the underlying connection binding them together.

This suggests two important differences between an agent who has achieved principled understanding and one who has not. First, principled understanding untethers one from one’s repertoire of memorized cases, since one need only rely on the principle going forward. Second, when faced with a new instance of the same underlying principle, the agent who understands that this novel case can be subsumed under that principle will be better placed to draw correct inferences than someone who merely has an unprincipled collection of previous instances to draw on. The achievement of principled understanding thus allows one’s competence to generalize beyond previously encountered cases.

In LLMs, we contend that principled understanding manifests when a model, instead of merely memorizing countless input-output pairs, discovers and implements an underlying algorithm or computational procedure that generates the correct outputs for an entire class of inputs. The evidence for this comes from the discovery of \textit{circuits} in LLMs: specific, self-contained subnetworks of attention heads and MLP layers that work together to perform a well-defined, reusable computation.

The discovery of such circuits suggests that LLMs can go beyond representing individual facts and learn the generative rules that connect them. This section explores the mechanistic evidence for this capacity, starting with a simple but foundational circuit (\ref{subsection:4.1}) before moving to an example of a model learning an abstract mathematical principle (\ref{subsection:4.2}). We then survey other findings to show that these are not isolated phenomena (\ref{subsection:4.3}) before considering the limitations that emerge once we distinguish “crystallized” from “fluid” understanding of principles (\ref{subsection:4.4}).

\subsection{A Simple Circuit: The Induction Head} \label{subsection:4.1}

A simple example of a circuit is the induction head, first identified in a two-layer toy model by \cite{elhage_mathematical_2021}. This circuit provides a mechanism for a basic but crucial form of in-context learning: completing repeating patterns. Specifically, it enables a model to solve tasks of the form “[A] [B] … [A] → [B].” This is an extremely useful pattern for natural language, as it allows the model to correctly complete a name like “Michael Jordan” by recognizing that this instance of “Michael” is the one that was previously followed by “Jordan” in the same text, rather than “Michael Jackson” or “Michael Faraday.”

The induction head circuit is an elegant two-step mechanism that relies on two attention heads at different layers, leveraging the tools we described in Section \ref{subsection:2.4}:

\begin{itemize}
    \item[] \textbf{Step 1: Marking the Prefix. }A first attention head---the “prefix head”---operates early in the model. When it processes token [B], it looks at the immediately preceding token [A]. Using the positional encoding that tells it token [A] is right before it, this head writes information into [B]’s residual stream that effectively marks it as “the token that comes after [A].”
    \item[] \textbf{Step 2: Retrieving the Completion.} Later in the model, a second attention head—the induction head proper—processes the \textit{second} instance of token [A]. Its query essentially asks: “Is there a token earlier in the text that was marked as ‘coming after a token like me’?” It scans the keys of all previous tokens and finds a strong match with the first token [B], which was marked by the prefix head. The induction head then copies the relevant information from [B]’s residual stream into the current stream, strongly upweighting the logit for [B] as the next token.\footnote{Remember from section 2 that the information gating is governed by key-query matching. The first attention head at layer one does this matching using positional encodings: it matches a query “I’m looking for position x-1” with the key “I’m at position x-1.” The second attention head at layer two matches the query “I’m looking for a token after an [A]” with the key “I’m after an [A].” This allows information to flow in the desired way. For a great visual explanation, we recommend \parencite{mcdougall_induction_2023}.}
\end{itemize}
\noindent
While this learned circuit still falls short of embodying a principle connecting facts, it constitutes an important first step towards it, because it implements a \textit{content}-\textit{agnostic} algorithm.\footnote{On the importance of non-content-specific representations in neural networks, see \cite{shea_moving_2021}.} The mechanism works just as well for completing “Harry Potter” as it does for “Marie Curie” or for repeating arbitrary sequences of random tokens. It is not learning a \textit{content}-\textit{specific} statistical association between “Michael” and “Jordan”; rather, it is implementing a \textit{general} \textit{procedure} for sequence completion by looking back in a targeted way.

The initial evidence for induction heads came from toy models. But subsequent behavioral analyses suggest that they are also a crucial component of in-context learning in large-scale models “in the wild” \parencite{olsson_-context_2022}.\footnote{They also study induction heads responsible for “fuzzy” copying tasks of the form [A*] [B*] … [A] → [B], such as in translation where [A*] and [B*] correspond to [A] and [B] in another language. Along those lines, Feucht et al. (\citeyear{feucht_dual-route_2025}) recently proposed and empirically investigated a distinction between \textit{token} induction heads and \textit{concept} induction heads.} While their evidence is primarily behavioral \citep[see][]{variengien_common_2023}, it clearly emerges that models are able to look back in the text in targeted ways to recall relevant token sequences. The induction head thus serves as a compelling first piece of evidence that LLMs can unify many specific instances of pattern completion under a single, learned circuit. 

\subsection{A Deeper Principle: The Fourier Algorithm for Modular Addition} \label{subsection:4.2}

While the induction head circuit does not connect facts or embody principled understanding, MI research has also found a striking example of a circuit that does seem to embody principled understanding. Nanda et al. (\citeyear{nanda_modular_addition_2023}) trained a simple one-layer transformer exclusively on examples of modular addition. Modular addition is addition on a circle or loop instead of a straight number line. This ensures that the numbers are always kept within a specific range, as they are with the hours on a clock or the days of the week. To perform modular addition, one adds the numbers normally before applying the modulus (the size of the loop): one calculates how many times the modulus fits into the sum of the numbers and takes the remainder. This remainder is the solution. When adding 9 + 5 with a modulus of 12, for example, one first calculates 9 + 5 = 14 before seeing how many times 12 goes into 14. Since 12 goes into 14 only once and leaves a remainder of 2, the solution to 9 + 5 (modulo 12) is 2. The model was trained on examples of modular addition using a modulus of 113 (e.g. 55 + 71 = 13), but was never given the rules of addition or the definition of the modulo operator. If LLMs were simply \textit{n}-gram models, one would expect the model to approach this task by simply memorizing the entire addition table given in the training data.

When the researchers trained the model, they found that it indeed first overfit the training set: accuracy on the training set soon converged to 100\%, but accuracy on the test set (the examples of modular addition that the model had not seen before) remained low. After around 10,000 passes through the training set, however, the model appeared to learn a mechanism that generalized, and its accuracy on the test set shot up to near 100\%.

This is the phenomenon of “grokking” we encountered in the introduction. Across different training runs, the researchers found that the models consistently exhibited grokking---but only when incentivized to rely on simple, smooth, and general functions. This was done by adding a regularization mechanism called “weight decay,” which introduces a selective pressure on weights to be close to zero. In addition to the usual penalty for deviating from the correct output, a model  incurs a penalty in proportion to the squared magnitude of its weights. This rewards the model for setting as many weights as possible as close to zero as possible while retaining accuracy, effectively pushing the model to form fewer, simpler, and more general computational circuits and discard information it no longer needs.

When they analyzed the model’s internal parameters and performed targeted interventions, the researchers discovered that the model had indeed gone beyond simply memorizing the addition table. It had learned to implement, within a compact computational circuit, a sophisticated general principle that allowed it to actually perform modular addition and calculate the answer. This is a transition from rote memorization to principled understanding: the model discovered the principle underlying all the facts of modular addition listed in the table.

To better understand the dynamics of this transition from memorization to the use of a “generalizing circuit,” the researchers employed an MI technique known as \textit{ablation}, which involves strategically deactivating specific parts of the network to observe the impact on performance. They conducted two key experiments at various points during training. First, they ablated the components identified as the generalizing circuit to measure how well the model performed when relying solely on memorization. Conversely, they ablated all \textit{other} components to isolate the generalizing circuit and assess its performance alone. This methodological approach allowed them to chart the development of both competing strategies and pinpoint when the generalizing circuit became fully formed and more effective than rote memorization.

The results suggest that grokking happens in three phases: a \textit{memorization} phase, where the network overfits the training data; a \textit{circuit} \textit{formation} phase, where the network learns a generalizing mechanism; and a \textit{cleanup} phase, where weight decay (the penalty encouraging the network to rely on small weights) removes memorized information. Interestingly, the transition to perfect test accuracy happened in the cleanup phase, after the generalizing mechanism was learned. The researchers took this to show that grokking is best understood as a gradual amplification of the generalizing circuit as memorized information is removed.

What principle did the network learn to implement? Rather than performing a modular addition by stacking number representations along a direction (like a slide rule\footnote{A \textit{slide} \textit{rule} is a simple mechanical calculator where numbers are marked on sliding rulers. To add, one ruler is physically shifted so that its zero aligns with a number on the fixed ruler, and the result is read directly from the new position—without breaking numbers into digits or carrying over.}), it found a smarter way: treating modular addition as a matter of rotating points on a circle. The key is to convert each number into an angle of rotation starting from a fixed point (e.g., the 3 o’clock position), with clockwise rotation representing positive numbers and counterclockwise rotation representing negative numbers.

Let us refer to the two numbers to be added as $a$ and $b$, the result as $c$, and the modulus as $P$. If we think of addition modulo $P$ in terms of rotation on a circle, a 360-degree rotation  corresponds to $P$. Therefore, each number \textit{a} will be an angle that represents the fraction $a/P$ of a full circle. For example, the number 0 corresponds to an angle of 0 degrees, placing it at 3 o’clock. For any angle, the model can easily compute the number’s $x$ and $y$ coordinates if it uses a circle with a radius of 1: the horizontal position ($x$-coordinate) is then given by the cosine of the angle, and the vertical position ($y$-coordinate) by the sine.

\begin{figure}[h!]
    \centering
    \includegraphics[width=\linewidth]{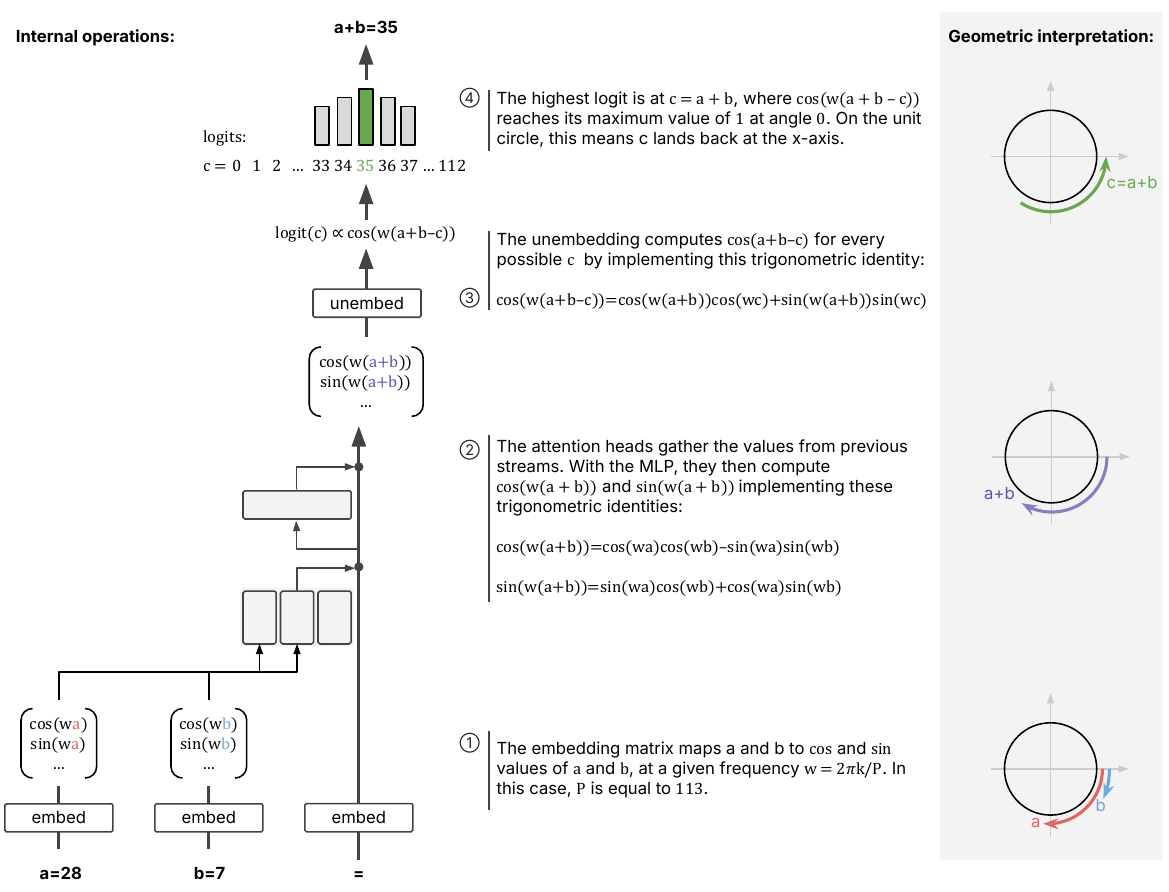}
     \vspace{-0.7cm}
    \caption{The Fourier multiplication algorithm. The single-layer transformer learns to perform modular addition by implementing the formulas for angle addition. Note that the cosine and sine components shown at different stages in parentheses are encoded as directions in latent space. The performed operations have a straightforward geometric interpretation: they do something like adding numbers on a clock. Note that the unembedding computes a logit for every possible $c$—checking how close each $c$ gets to the value of $a + b$ (see also Fig. \ref{fig:modular2}b). On the “clock,” this means winding back for every possible $c$ value starting from $a + b$. The maximal logit will be obtained when $c = a + b$, corresponding to the $c$ value that gets back to the starting point (where the angle is zero and $\cos$ is maximized).}
    \label{fig:modular1}
\end{figure}

Having mapped the numbers $a$ and $b$ onto a circle, the model’s next task is to perform addition by combining their rotations. Geometrically, this is simple: rotate by angle $a$, then rotate again by angle $b$ to find the final position. The model learns to perform the algebraic equivalent of this operation by implementing the trigonometric angle addition formulas (Fig. \ref{fig:modular1}). These formulas provide a direct computational path from the four known coordinates (the sine and cosine of $a$ and $b$) to the two desired coordinates (the sine and cosine of $a + b$). The angle addition rules are nothing more than a specific pattern of multiplications and additions, which the model can encode directly into the weights of its MLP layers. It effectively learns a circuit for doing modular arithmetic by adding angles.

\begin{figure}[h!]
    \centering
    \includegraphics[width=\linewidth]{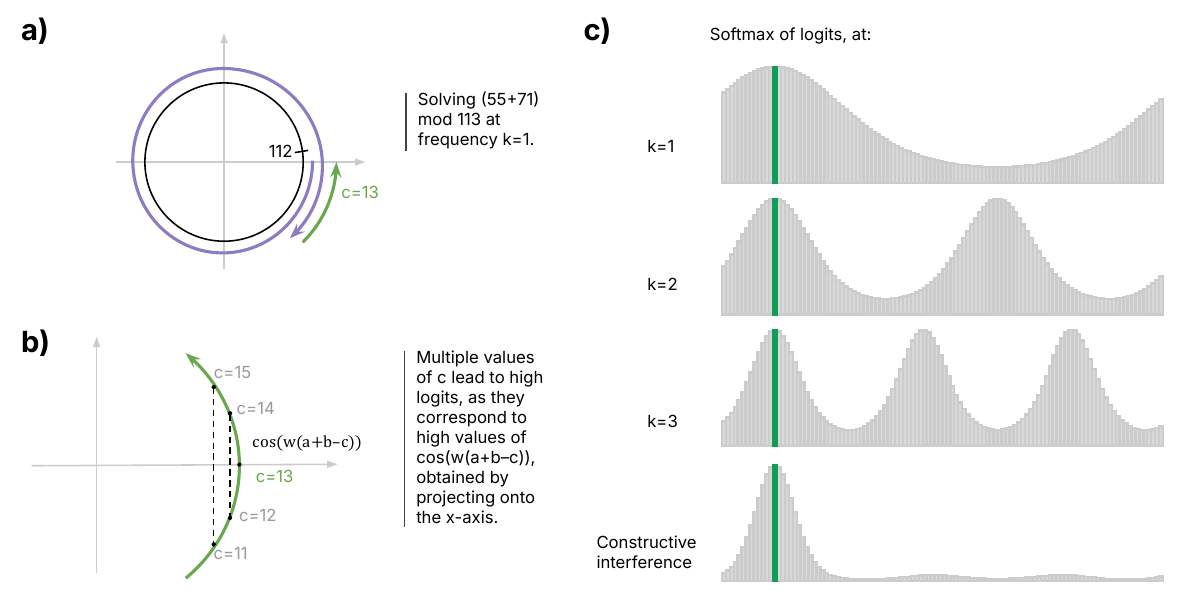}
     \vspace{-0.7cm}
    \caption{The two advantages of the Fourier multiplication algorithm: the circle-wrapping trick and the constructive interference trick. (a) The circle-wrapping trick exploits the periodicity of the cosine function. When performing 55 + 71, the model adds the corresponding angles on the unit circle. This results in a total rotation of 126 steps. The angle wraps around the circle once, and, when it passes step 112, resets to zero—finally landing at position $c = 13$. (b) However, because the cosine function is relatively flat near its maximum, neighboring values of $c$ also receive high logit scores, making the algorithm less precise. At frequency $k = 1$, the cosine peaks at the correct value (high accuracy) but has a broad profile (low precision). (c) As the frequency $k$ increases (e.g., $k = 2$ or $k = 3$), the angle wraps around the unit circle $k$ times faster. This creates $k$ distinct peaks: while each peak becomes sharper (indicating higher precision), the model becomes less accurate overall, since it now has multiple plausible candidate values for the answer. When $k > 1$, the model is effectively performing the addition modulo $P / k$ rather than modulo $P$. For example, at $k = 2$, the model not only peaks at the correct value (e.g., $c = 13$), but also at the value $P / 2$ (e.g., if $P = 113$, the second peak appears around $113 / 2$). By combining signals from multiple frequencies, the model achieves $\textit{constructive interference}$, reinforcing the correct output while canceling out others---thus achieving both high precision and high accuracy.}
    \label{fig:modular2}
\end{figure}

The power of this solution lies in how it exploits the periodicity of the sine and cosine functions, which provide a perfect computational scaffold for the "modulo" part of the operation.\footnote{For a clear interactive visualization, see https://www.mathsisfun.com/algebra/trig-interactive-unit-circle.html.} Periodicity simply means that the functions repeat their values in a regular cycle. Think of a clock: after 12 o'clock, the next hour is not 13, but 1—the sequence wraps around and repeats. Sine and cosine behave the same way. The cosine of a 370-degree angle is identical to the cosine of a 10-degree angle, because the rotation has simply “wrapped around” the circle. For the model, this means that it does not need to learn the abstract rule of “see how often $P$ goes into $a + b$ and take the remainder.” When adding 55 and 71 (modulo 113), the model rotates 55 steps from zero and then another 71 steps, naturally landing on the same point on the circle as the angle for 13 (Fig. \ref{fig:modular2}a). The model gets the correct “wrapped around” result automatically, through the inherent properties of the geometric system it has learned to use. This \textit{circle-wrapping trick} allows the model to perform modular addition by simply adding the angles of the numbers.

However, this method also has a weakness: the cosine curve is relatively flat near its peak (Fig. \ref{fig:modular2}b). This means that while the correct answer $c = 13$ gets the highest score, nearby values like 12 and 14 also get high scores, making the algorithm imprecise. To remedy this, the model brilliantly employs a second trick: the \textit{constructive interference trick} (Fig. \ref{fig:modular2}c). It performs the same clock-like addition simultaneously on multiple circles, each spinning at a different frequency $k$. A higher frequency means the angle wraps around the circle $k$ times faster. On a high-frequency circle, the cosine peaks are much sharper (higher precision), but there are multiple peaks, making it inaccurate (e.g., for $k = 2$, it peaks at the correct answer $c = 13$, but also at $c = 13 + 113 / 2$). By combining the signals from five different frequencies, the model ensures that only the single, correct answer receives a high score across every circle: the incorrect peaks from the high-frequency circles are cancelled out by the low scores they receive on other circles. This constructive interference allows the model to achieve both high accuracy and high precision.

This method is named the “Fourier multiplication algorithm” for two reasons. First, its use of sine and cosine components across multiple frequencies is the hallmark of Fourier analysis. Second, the “multiplication” refers to how the model achieves the equivalent of addition by implementing the angle addition formulas, which are fundamentally based on multiplying the sine and cosine of the input angles.

In essence, the model managed to move from a vast collection of disconnected facts (the examples of correct modular addition it had been given, such as “55 + 71 = 13”) to the principle or rule that connects them all and that also underlies the examples of modular addition that were not in the training set. Furthermore, the model even discovered that by representing numbers on circles, it could transform a complex, two-step problem (add, then apply the modulus) into a single-step geometric problem (rotate on circles) that was a perfect fit for its own architecture. Instead of storing countless individual examples of modular addition, the model compressed these into a computational circuit implementing the principle underlying the panoply of examples. This is not only more representationally efficient, allowing the model to reduce the weight decay penalty, but also allows it to generate the correct answer for all possible inputs rather than merely for the training set. This feat of compressing a variety of initially disconnected facts down to the underlying principle that connects and unifies them, thereby enabling the model to generalize, strikingly displays the hallmarks of principled understanding.

This simple circuit functionally satisfies the main conditions for explanatory understanding discussed in contemporary epistemology (see Baumberger et al.,  \citeyear{baumberger2017understanding}). It embodies a network of explanatory connections \parencite{khalifa2017scientific, Strevens2013-STRNUW}, which satisfy the condition of factivity \parencite{kvanvig2009value, Kelp2015}, since they rely on geometric facts (the trigonometric identities) that actually obtain. These connections enable the unification of many mathematical facts under a single principle \parencite{friedman1974explanation, kitcher1981unification}, which in turn allows the model to compress its repository of memorized cases down to a single circuit implementing this principle \parencite{wilkenfeld2018compression}. This also bestows upon the model the inferential capacities associated with grasping \parencite{hills2015understanding, grimm2011understanding}—manifested here in the model’s ability to generalize correctly to all inputs, not just those in the training set. Finally, the circuit satisfies multiple criteria linked to the condition of justification: it has a wide scope, is used consistently and reflects a pursuit of simplicity via weight decay \parencite{baumberger2017understanding, Beckmann2025}.

\subsection{Circuits in the Wild} \label{subsection:4.3}

While the modular addition circuit was found in a toy model, a growing body of research is uncovering similarly complex circuits in state-of-the-art LLMs. Research on regular (non-modular) addition “in the wild” suggests that real LLMs also rely on operations across multiple frequencies—or, equivalently, across multiple moduli \parencite{zhou_pre-trained_2024, ameisen_circuit_2025, kantamneni_language_2025, nikankin_arithmetic_2025}. Other examples of circuits in the wild include an indirect object identification circuit \parencite{wang_interpretability_2023}, a greater-than circuit for numerical comparison \parencite{hanna_how_2023}, and a multiple-choice question circuit \parencite{lieberum_does_2023}.

Recently, researchers at Anthropic have developed techniques to automatically extract functional circuits for specific prompts from frontier models like Claude 3.5 Sonnet \parencite{ameisen_circuit_2025, lindsey_biology_2025}. They have successfully identified circuits responsible for tasks as diverse as entity recognition, planning the structure of a poem, generating medical diagnoses, and even refusing to answer unsafe queries. A particularly elegant example is a line-breaking circuit in Claude 3.5 Haiku that uses geometric transformations—rotating representations of character count relative to line length in its attention blocks—to detect when the model is approaching the end of a line \parencite{gurnee2025when}. These findings confirm that the principle of computation via specialized circuits is not an artifact of toy models, but a core operating principle of modern LLMs.

In summary, the mechanistic evidence for circuits provides strong support for what we have termed principled understanding. These circuits demonstrate that LLMs can move beyond learning correlational patterns to implementing general, reusable algorithms. By discovering these underlying principles, the model achieves a powerful form of unification, allowing it to generalize far more effectively and efficiently than if it were merely memorizing its training data. This capacity represents the highest level of understanding in our proposed hierarchy, revealing a computational sophistication in LLMs that a purely statistical view fails to capture.

\subsection{Crystallized vs. Fluid Understanding of Principles}
\label{subsection:4.4}
The modular addition circuit discussed above represents a significant cognitive achievement: the model has successfully distilled a general algorithm from specific data points. However, this form of principled understanding remains in an important sense \textit{static}. Once training ends, the circuit that has emerged is frozen. Consequently, the understanding that crystallized during training is now available at inference, but it does not continue to adapt to new challenges. Can LLMs also identify a novel underlying principle and implement a corresponding algorithm on the fly, when faced with a new type of problem? In other words, do they exhibit \textit{fluid} as well as crystallized understanding?
	
This distinction mirrors the established contrast in psychology between \textit{crystallized} intelligence—the application of learned knowledge and practiced skill—and \textit{fluid} intelligence—the capacity to solve novel problems without leaning much on specific prior learning. While the modular addition circuit offers an impressive example of crystallized understanding, LLMs’ capacity for fluid understanding is a separate question.
	
The primary testing ground for this capability is the Abstraction and Reasoning Corpus (ARC-AGI), which was expressly designed to test a system’s ability to discover the rules or principles underlying something at inference time, i.e. after training has been completed \parencite{chollet2019measureintelligence, chollet2025arcprize2024technical}. To succeed on ARC, a model must be able to look at a few pairs of inputs and outputs, infer the abstract rule by which one is transformed into the other (e.g., “gravity pulls all blue pixels down until they hit a red pixel”), and apply that rule to a test input.
	
While humans find these tasks intuitive, LLMs typically have a harder time discovering principles in real time.\footnote{A related limitation emerges from recent investigations into LLMs’ maze-solving capabilities \parencite{palod2025performativethinkingbrittlecorrelation, valmeekam2025semanticsunreasonableeffectivenessreasonless}.} This tendency to default to static, training-induced mechanisms has led critics to argue that LLMs lack true intelligence, reducing them to combinators of “heuristics” \parencite{mitchell_llms_2025}, “program templates” \parencite{chollet2024recombination}, or pieces of “procedural knowledge” \parencite{kambhampati2025memory}.
 
These criticisms should not be taken to preclude all forms of understanding, however. While they correctly highlight limitations of LLMs’ \textit{fluid} understanding, training-induced mechanisms can still amount to \textit{crystallized} understanding. As seen with modular addition, a “program template” or “procedure” can represent a grasp of structure that is sufficient to qualify as a form of understanding.

\section{A Motley Mix of Mechanisms} \label{section:5}

So far, we have presented mechanistic evidence that LLMs are \textit{capable} of forms of understanding comparable to our own. They can identify entities and recall associated information. They can play board games by constructing internal models of the board state without visual input. And, to perform arithmetic, they can transition from rote memorization to the application of underlying principles.

But does the existence of these capacities mean that LLMs \textit{actually} understand in typical cases? Zooming out from the most compelling, fine-grained findings of mechanistic interpretability to date reveals a more complex reality. Insofar as LLMs instantiate forms of cognition, they instantiate highly \textit{strange} forms of it, shaped by constraints and incentives that differ starkly from those shaping human intelligence \parencite{fleuret_strange_2023, shanahan_murray_simulacra_2024}. This divergence complicates the attribution of understanding.

Perhaps most notably, the architecture, memorization capacity, and training objectives of LLMs place little constraint on the sheer \textit{quantity} of mechanisms they develop in response to training. As a result, LLMs end up stocked not with a single mechanism for each task, but with an entire assemblage of mechanisms. 

At inference, these mechanisms operate \textit{in parallel}: LLMs typically tackle problems by deploying
multiple distinct mechanisms simultaneously rather than relying on a single circuit \parencite{lindsey_biology_2025}. These mechanisms may cooperate, compete, or contribute independent pieces, with the final output emerging from their combined effect. Higher-quality circuits are sometimes reinforced, but sometimes also drowned out by lower-quality circuits.

The constructive interference observed in modular addition offers a glimpse of this: the model’s final answer arises from the aggregation of five separate frequency-based calculations. Similarly, internal circuitry for standard addition exemplifies the deployment of numerous independent heuristics, such as multiple modulo mechanisms---e.g., computing a + b modulo 2 to determine if the sum is even or odd, and modulo 10 to determine its final digit \parencite{zhou_pre-trained_2024, ameisen_circuit_2025, kantamneni_language_2025, nikankin_arithmetic_2025}. These are supplemented by parallel range-finding mechanisms of varying precision, which determine the approximate magnitude of the sum \parencite{ameisen_circuit_2025, nikankin_arithmetic_2025}. The final answer emerges from the collective consensus of this “bag of tricks.”\footnote{To compute 36 + 59, the LLM Claude 3.5 relies on the additive combination of a range-finding mechanism placing the sum at $\sim92$; a modulo 100 mechanism according to which the sum ends with 95; and a modulo 10 mechanism according to which the sum ends with 5. Together, these mechanisms constructively interfere at 36 + 59 = 95 \parencite{ameisen_circuit_2025}. Interestingly, the modulo mechanisms resemble how humans carry over digits when adding—effectively adding in steps of modulo 10, 100, 1000, and so on.}

Othello-GPT likewise deploys a mix of mechanisms in parallel: localized decision rules—like one that updates specific squares only when A4 is played with certain neighbors occupied; specific board pattern detectors that look for states that make moves legal; and timing-based heuristics that rely on “clock neurons” to track the game’s progress and upweight moves that are statistically more likely in the late game, such as those targeting edge squares \parencite{lin_othellogpt_2024}.

Even factual recall relies on distinct pathways operating in parallel. When prompted with “Michael Jordan plays the sport of,” Claude 3.5 uses at least two pathways. One pathway, activated by words like “plays” and “sport,” broadly boosts sports-related features, including both “basketball” and “football.” Simultaneously, a second pathway, triggered by “Michael Jordan,” strongly activates “basketball” while suppressing alternatives. The correct output—“basketball”—emerges from the confluence of these separate streams \parencite{ameisen_circuit_2025, chughtai_summing_2024}.

The strangeness of LLM cognition is compounded by the drastic variation in the \textit{quality} of these mechanisms. They form a “motley mix” spanning the full hierarchy of mechanisms: from pure Blockhead-style memorization of training data \parencite{carlini2023quantifying, aerni2024measuring, stoehr2024localizing} through shortcuts that rely on spurious correlations \parencite{geirhos2020shortcut} and shallow heuristics that rely on superficial statistics (e.g., Ma et al., \citeyear{ma2025memorization}) to mechanisms encoding the kinds of conceptual, state-of-the-world, and principled understanding we have detailed.

Beyond examining particular examples, this motley mix phenomenon can be established systematically through loss curvature analysis \parencite{merullo2025memorizationreasoningspectrumloss}. This approach asks a simple question: which parts of a model are such that even small weight changes cause large increases in average loss across different inputs, and which parts are such that even large changes barely affect average loss? This can be measured by examining the curvature of the loss landscape with respect to weights, averaged across many inputs. Weights implementing mechanisms recruited across many inputs reinforce the increase in loss and appear as high-curvature; weights recruited only in specific cases produce sharp increases in loss only in those cases, which wash out when averaged, registering as low-curvature or “flat.”

Merullo et al. confirmed that model components vary drastically in their loss curvature, validating the motley mix picture. Removing low-curvature components suppressed verbatim recitation and closed-book question-answering while largely preserving more general capacities, such as logical reasoning and open-book question-answering. On Merullo et al.’s interpretation, this amounts to disentangling the weights underwriting “shared, reusable structure” from those supporting “recitation and brittle behaviors.”

An LLM is thus a \textit{motley mix of mechanisms}: a system where many circuits of varying quality crystallize during training and operate in parallel at inference. These mechanisms occasionally interact via activation or inhibition, but seem to lack centralized or hierarchical control.\footnote{Adding a form of hierarchical control to transformer-based models is currently a hot topic in AI \parencite{jolicoeur-martineau2025less, behrouz2025nested}. Note that the meta-cognitive features discussed in subsection \ref{subsection:2.1} represent early signs of hierarchical processing, but function more as scattered and isolated checks.} LLM outputs accordingly resemble the verdict of a gigantic committee of drastically varying expertise, working largely in isolation with only occasional collaboration. 

The fact that LLMs function more like a heterogeneous committee than like a single, unified mind  complicates the attribution of understanding. When LLMs rely purely on memorized answers or shallow shortcuts, they clearly fail to understand. Given their immense capacity to store such “cheap” mechanisms, these failures are common; behavior that warrants the attribution of understanding in humans may not do so in LLMs.

The most philosophically complex cases arise, however, when LLMs deploy conceptual and principled unifications---the very mechanisms that warrant attributing understanding—yet combine them with cheaper, shallower heuristics. Often, the model achieves generalization only through the collective interference of this mixed bag.

One might argue that this reliance on heuristics should not in itself be disqualifying. After all, cognitive psychology confirms that humans themselves rely heavily on shortcuts and parallel processing across competing mechanisms (Kahneman, \citeyear{kahneman2011thinking}; Sodium \citeyear{sodium2024heuristics}; Summerfield, \citeyear{summerfield2025these}, Chap. 20).  

Yet humans possess a distinctive meta-cognitive ability to override heuristics and unify phenomena using increasingly parsimonious explanatory structures. This striving for parsimonious systematicity is widely regarded as a hallmark of world understanding \parencite{queloz2025a, queloz2025b}. As Einstein put it, the “grand aim of all science ... is to cover the greatest possible number of empirical facts by logical deductions from the smallest possible number of hypotheses or axioms” (quoted in Nash \citeyear{nash1963}, p. 173). It is this mature understanding—marked by a striving for parsimony, where a few principles unify a wide range of phenomena—that supports \textit{robust} explanation and prediction and ultimately warrants epistemic trust \parencite{deRegt2009-DERTEV, QuelozManuscript-QUEWWC-2}.

The \textit{motley mix} phenomenon, however, undermines this basis for epistemic trust: even broad behavioral success cannot reveal whether the model relies on parsimonious unifying mechanisms or merely a swarm of shallow heuristics. Moreover, even when researchers do identify circuits implementing principled, parsimonious understanding, these understanding-bearing mechanisms might still be overwhelmed by other heuristics. Researchers observed this problem in syllogistic inference: they identified a formal circuit that correctly processes logical structure, but gets drowned out by content-based mechanisms \parencite{kim2025reasoningcircuitslanguagemodels, valentino2025mitigatingcontenteffectsreasoning}. For instance, the model is less likely to accept the syllogism: “All apples are vegetations. All vegetations are institutions. Some apples are institutions” than “All apples are edible fruits. All edible fruits are fruits. All apples are fruits”—despite both being logically valid—simply because the first contains substantively implausible premises that triggers content-based circuits. Because whatever genuine understanding an LLM possesses is buried within a sprawl of shallower heuristics, it cannot ground the same epistemic trust we grant to more parsimonious minds.

\section{Conclusion} \label{section:conclusion}

The inner world of an LLM is thus not the statistical void implied by the deflationary view. MI provides compelling evidence that LLMs develop internal structures that are functionally analogous to core aspects of human understanding---specifically, the kinds of understanding that consist in seeing connections. These include the connections between different manifestations of entities or properties that we draw when we subsume them under a single concept; the connections between concepts that we draw when we learn about the factual state of the world; and the connections between facts that we draw when we break through to an underlying principle. We conclude that talk of “understanding” in LLMs can be made precise and empirically tractable.

Yet this investigation has also revealed a profound alienness. LLMs do not strive for parsimony as humans do; whatever understanding they possess is distributed across a motley mix of mechanisms. Recognizing this need not undermine the claim that LLMs understand, but it does require us to draw finer distinctions between varieties of understanding and to forge conceptions that are decoupled from the specificities of human understanding.\footnote{For discussions of this need for new ways of thinking and talking about LLMs, see Shanahan \citeyear{shanahantalking1, shanahantalking2} and Müller \& Löhr \citeyear{muellerloehr}.}

We have argued that combining philosophy and MI allows us to move beyond the binary question of \textit{whether} LLMs understand. It facilitates a more fine-grained project: identifying the specific kinds of understanding LLMs realize; mapping their characteristic failure modes and divergences from human cognition; and drawing out the implications for epistemic trust. These new forms of understanding---at once transparent to inspection and alien in operation---thus open up the prospect of a new, mechanistically grounded comparative epistemology.

\clearpage

\bibliographystyle{apalike}
\bibliography{refs}

\newcommand{\noop}[1]{}
\begin{thebibliography}{}

\bibitem[Aerni et~al., 2024]{aerni2024measuring}
Aerni, M., Rando, J., Debenedetti, E., Carlini, N., Ippolito, D., and Tram\`er, F. (2024).
\newblock Measuring non-adversarial reproduction of training data in large language models.
\newblock {\em arXiv preprint arXiv:2411.10242}.

\bibitem[Alain and Bengio, 2017]{alain_understanding_2017}
Alain, G. and Bengio, Y. (2017).
\newblock Understanding intermediate layers using linear classifier probes.
\newblock In {\em 5th {International} {Conference} on {Learning} {Representations}, {ICLR} 2017, {Toulon}, {France}, {April} 24-26, 2017, {Workshop} {Track} {Proceedings}}. OpenReview.net.

\bibitem[Ameisen et~al., 2025]{ameisen_circuit_2025}
Ameisen, E., Lindsey, J., Pearce, A., Gurnee, W., Turner, N.~L., Chen, B., Citro, C., Abrahams, D., Carter, S., Hosmer, B., Marcus, J., Sklar, M., Templeton, A., Bricken, T., McDougall, C., Cunningham, H., Henighan, T., Jermyn, A., Jones, A., Persic, A., Qi, Z., Ben~Thompson, T., Zimmerman, S., Rivoire, K., Conerly, T., Olah, C., and Batson, J. (2025).
\newblock Circuit {Tracing}: {Revealing} {Computational} {Graphs} in {Language} {Models}.
\newblock {\em Transformer Circuits Thread}.

\bibitem[{Anthropic}, 2024]{anthropic_golden_2024}
{Anthropic} (2024).
\newblock Golden {Gate} {Claude}.

\bibitem[Arditi et~al., 2024]{arditi_refusal_2024}
Arditi, A., Obeso, O., Syed, A., Paleka, D., Panickssery, N., Gurnee, W., and Nanda, N. (2024).
\newblock Refusal in {Language} {Models} {Is} {Mediated} by a {Single} {Direction}.
\newblock \_eprint: 2406.11717.

\bibitem[Ayonrinde and Jaburi, 2025]{Ayonrinde_Jaburi_2025}
Ayonrinde, K. and Jaburi, L. (2025).
\newblock A mathematical philosophy of explanations in mechanistic interpretability.
\newblock {\em Proceedings of the AAAI/ACM Conference on AI, Ethics, and Society}, 8(1):265--278.

\bibitem[Baumberger et~al., 2017]{baumberger2017understanding}
Baumberger, C., Beisbart, C., and Brun, G. (2017).
\newblock What is understanding? an overview of recent debates in epistemology and philosophy of science.
\newblock In {\em Explaining Understanding: New Perspectives from Epistemology and Philosophy of Science}, pages 1--34.

\bibitem[Bayat et~al., 2025]{bayat_steering_2025}
Bayat, R., Rahimi-Kalahroudi, A., Pezeshki, M., Chandar, S., and Vincent, P. (2025).
\newblock Steering {Large} {Language} {Model} {Activations} in {Sparse} {Spaces}.
\newblock \_eprint: 2503.00177.

\bibitem[Beckmann, 2025]{Beckmann2025}
Beckmann, P. (2025).
\newblock New horizons in machine understanding: explanatory and objectual understanding in deep learning video generation models.
\newblock {\em Synthese}, 206:285.

\bibitem[Beckmann, 2026]{BeckmannManuscript-BECDLM-2}
Beckmann, P. (2026).
\newblock Deep learning models also recall features.

\bibitem[Behrouz et~al., 2025]{behrouz2025nested}
Behrouz, A., Razaviyayn, M., Zhong, P., and Mirrokni, V. (2025).
\newblock Nested learning: The illusion of deep learning architectures.
\newblock In {\em Proceedings of the Thirty‑Ninth International Conference on Neural Information Processing Systems (NeurIPS 2025)}. NeurIPS.

\bibitem[Beisbart, 2025]{beisbart2025}
Beisbart, C. (2025).
\newblock Digging deeper with deep learning? explanatory understanding and deep neural networks.
\newblock {\em European Journal for Philosophy of Science}, 15(3):40.

\bibitem[Belkoniene, 2023]{belkoniene2023}
Belkoniene, M. (2023).
\newblock Grasping in understanding.
\newblock {\em The British Journal for the Philosophy of Science}, 74(3):603--617.
\newblock doi: 10.1086/714816.

\bibitem[Bender et~al., 2021]{bender2021dangers}
Bender, E.~M., Gebru, T., McMillan-Major, A., and Shmitchell, S. (2021).
\newblock On the dangers of stochastic parrots: Can language models be too big?
\newblock In {\em Proceedings of the 2021 ACM conference on fairness, accountability, and transparency}, pages 610--623.

\bibitem[Bender and Koller, 2020]{bender2020climbing}
Bender, E.~M. and Koller, A. (2020).
\newblock Climbing towards nlu: On meaning, form, and understanding in the age of data.
\newblock In {\em Proceedings of the 58th annual meeting of the association for computational linguistics}, pages 5185--5198.

\bibitem[Bengio et~al., 2013]{bengio_representation_2013}
Bengio, Y., Courville, A., and Vincent, P. (2013).
\newblock Representation {Learning}: {A} {Review} and {New} {Perspectives}.
\newblock {\em IEEE Trans. Pattern Anal. Mach. Intell.}, 35(8):1798--1828.
\newblock Place: USA Publisher: IEEE Computer Society.

\bibitem[Block, 1981]{block1981psychologism}
Block, N. (1981).
\newblock Psychologism and behaviorism.
\newblock {\em Philosophical Review}, 90(1):5--43.

\bibitem[Block, 1987]{block1987functional}
Block, N. (1987).
\newblock Functional role and truth conditions.
\newblock {\em Proceedings of the Aristotelian Society, Supplementary Volume}, 61:157--181.

\bibitem[Bricken et~al., 2023]{bricken_towards_2023}
Bricken, T., Templeton, A., Batson, J., Chen, B., Jermyn, A., Conerly, T., Turner, N., Anil, C., Denison, C., Askell, A., Lasenby, R., Wu, Y., Kravec, S., Schiefer, N., Maxwell, T., Joseph, N., Hatfield-Dodds, Z., Tamkin, A., Nguyen, K., McLean, B., Burke, J.~E., Hume, T., Carter, S., Henighan, T., and Olah, C. (2023).
\newblock Towards {Monosemanticity}: {Decomposing} {Language} {Models} {With} {Dictionary} {Learning}.
\newblock {\em Transformer Circuits Thread}.

\bibitem[Brown et~al., 2020]{brown2020language}
Brown, T., Mann, B., Ryder, N., Subbiah, M., Kaplan, J.~D., Dhariwal, P., Neelakantan, A., Shyam, P., Sastry, G., Askell, A., et~al. (2020).
\newblock Language models are few-shot learners.
\newblock {\em Advances in Neural Information Processing Systems}, 33:1877--1901.

\bibitem[Bussmann et~al., 2025]{bussmann_learning_2025}
Bussmann, B., Nabeshima, N., Karvonen, A., and Nanda, N. (2025).
\newblock Learning {Multi}-{Level} {Features} with {Matryoshka} {Sparse} {Autoencoders}.
\newblock \_eprint: 2503.17547.

\bibitem[Carlini et~al., 2023]{carlini2023quantifying}
Carlini, N., Ippolito, D., Jagielski, M., Lee, K., Tr\'amer, F., and Zhang, C. (2023).
\newblock Quantifying memorization across neural language models.
\newblock In {\em Proceedings of the Eleventh International Conference on Learning Representations}, City, Country -- ICLR (virtual/poster session).

\bibitem[Chang, 2025]{chang2025n}
Chang, T.~A. (2025).
\newblock {\em N-Gram Learning and Pretraining Dynamics in Transformer Language Models}.
\newblock PhD thesis, University of California, San Diego.

\bibitem[Chanin et~al., 2024]{chanin_is_2024}
Chanin, D., Wilken-Smith, J., Dulka, T., Bhatnagar, H., and Bloom, J. (2024).
\newblock A is for {Absorption}: {Studying} {Feature} {Splitting} and {Absorption} in {Sparse} {Autoencoders}.
\newblock \_eprint: 2409.14507.

\bibitem[Chollet, 2019]{chollet2019measureintelligence}
Chollet, F. (2019).
\newblock On the measure of intelligence.

\bibitem[Chollet, 2024]{chollet2024recombination}
Chollet, F. (2024).
\newblock Note that there's a continuous spectrum between reusing program templates you've learned, and actual reasoning.
\newblock X (formerly Twitter) post. Available at \url{https://x.com/fchollet/status/1800348890094563494}.

\bibitem[Chollet et~al., 2025]{chollet2025arcprize2024technical}
Chollet, F., Knoop, M., Kamradt, G., and Landers, B. (2025).
\newblock Arc prize 2024: Technical report.

\bibitem[Chughtai et~al., 2024]{chughtai_summing_2024}
Chughtai, B., Cooney, A., and Nanda, N. (2024).
\newblock Summing {Up} the {Facts}: {Additive} {Mechanisms} behind {Factual} {Recall} in {LLMs}.

\bibitem[de~Regt, 2009]{deRegt2009-DERTEV}
de~Regt, H.~W. (2009).
\newblock The epistemic value of understanding.
\newblock {\em Philosophy of Science}, 76(5):585--597.

\bibitem[De~Regt, 2017]{de_regt2017scientific}
De~Regt, H.~W. (2017).
\newblock {\em Understanding scientific understanding}.
\newblock Oxford University Press.

\bibitem[DeMoss et~al., 2024]{demoss2024complexity}
DeMoss, B., Sapora, S., Foerster, J., Hawes, N., and Posner, I. (2024).
\newblock The complexity dynamics of grokking.
\newblock {\em arXiv preprint arXiv:2412.09810}.

\bibitem[Dewey, 1929]{Dewey1929-DEWTQF}
Dewey, J. (1929).
\newblock {\em The Quest for Certainty: A Study of the Relation of Knowledge and Action}.
\newblock Putnam, New York,.

\bibitem[Du et~al., 2023]{du2024shortcutllm}
Du, M., He, F., Zou, N., Tao, D., and Hu, X. (2023).
\newblock Shortcut learning of large language models in natural language understanding.
\newblock {\em Commun. ACM}, 67(1):110–120.

\bibitem[Elhage et~al., 2022]{elhage_toy_2022}
Elhage, N., Hume, T., Olsson, C., Schiefer, N., Henighan, T., Kravec, S., Hatfield-Dodds, Z., Lasenby, R., Drain, D., Chen, C., Grosse, R., McCandlish, S., Kaplan, J., Amodei, D., Wattenberg, M., and Olah, C. (2022).
\newblock Toy {Models} of {Superposition}.
\newblock {\em Transformer Circuits Thread}.

\bibitem[Elhage et~al., 2021]{elhage_mathematical_2021}
Elhage, N., Nanda, N., Olsson, C., Henighan, T., Joseph, N., Mann, B., Askell, A., Bai, Y., Chen, A., Conerly, T., DasSarma, N., Drain, D., Ganguli, D., Hatfield-Dodds, Z., Hernandez, D., Jones, A., Kernion, J., Lovitt, L., Ndousse, K., Amodei, D., Brown, T., Clark, J., Kaplan, J., McCandlish, S., and Olah, C. (2021).
\newblock A {Mathematical} {Framework} for {Transformer} {Circuits}.
\newblock {\em Transformer Circuits Thread}.

\bibitem[Ferrando et~al., 2025]{ferrando_i_2025}
Ferrando, J., Obeso, O.~B., Rajamanoharan, S., and Nanda, N. (2025).
\newblock Do {I} {Know} {This} {Entity}? {Knowledge} {Awareness} and {Hallucinations} in {Language} {Models}.
\newblock In {\em The {Thirteenth} {International} {Conference} on {Learning} {Representations}}.

\bibitem[Feucht et~al., 2025]{feucht_dual-route_2025}
Feucht, S., Todd, E., Wallace, B., and Bau, D. (2025).
\newblock The {Dual}-{Route} {Model} of {Induction}.
\newblock \_eprint: 2504.03022.

\bibitem[Field, 1977]{field1977logic}
Field, H.~H. (1977).
\newblock Logic, meaning, and conceptual role.
\newblock {\em The Journal of Philosophy}, 74(7):379--409.

\bibitem[Fleuret, 2023]{fleuret_strange_2023}
Fleuret, F. (2023).
\newblock Strange {Minds}.

\bibitem[Friedman, 1974]{friedman1974explanation}
Friedman, M. (1974).
\newblock Explanation and scientific understanding.
\newblock {\em The Journal of Philosophy}, 71(1):5--19.

\bibitem[Geirhos et~al., 2020]{geirhos2020shortcut}
Geirhos, R., Jacobsen, J.-H., Michaelis, C., Zemel, R., Brendel, W., Bethge, M., and Wichmann, F.~A. (2020).
\newblock Shortcut learning in deep neural networks.
\newblock In {\em Nature Machine Intelligence}, volume~2, pages 665--673. Nature Publishing Group.

\bibitem[{Gemini Team} et~al., 2025]{geminiteam2025geminifamilyhighlycapable}
{Gemini Team}, Anil, R., Borgeaud, S., Alayrac, J.-B., Dean, J., and Vinyals, O. (2025).
\newblock Gemini: A family of highly capable multimodal models.

\bibitem[Geva et~al., 2023]{geva_dissecting_2023}
Geva, M., Bastings, J., Filippova, K., and Globerson, A. (2023).
\newblock Dissecting {Recall} of {Factual} {Associations} in {Auto}-{Regressive} {Language} {Models}.
\newblock In Bouamor, H., Pino, J., and Bali, K., editors, {\em Proceedings of the 2023 {Conference} on {Empirical} {Methods} in {Natural} {Language} {Processing}}, pages 12216--12235, Singapore. Association for Computational Linguistics.

\bibitem[Ginsborg, 2018]{ginsborg2018normativity}
Ginsborg, H. (2018).
\newblock Normativity and concepts.
\newblock In Star, D., editor, {\em The Oxford Handbook of Reasons and Normativity}, pages 989--1014. Oxford University Press, New York.

\bibitem[Goldstein and Levinstein, 2024]{goldstein2024}
Goldstein, S. and Levinstein, B.~A. (2024).
\newblock Does chatgpt have a mind?
\newblock {\em arXiv preprint arXiv:2407.11015}.

\bibitem[Grimm, 2006]{grimm2006species}
Grimm, S. (2006).
\newblock Is understanding a species of knowledge?
\newblock {\em The British Journal for the Philosophy of Science}.

\bibitem[Grimm, 2011]{grimm2011understanding}
Grimm, S. (2011).
\newblock Understanding.
\newblock In {\em The Routledge companion to epistemology}, pages 84--93. Routledge.

\bibitem[Grimm, 2025]{sep-understanding}
Grimm, S. (2025).
\newblock {Understanding}.
\newblock In Zalta, E.~N. and Nodelman, U., editors, {\em The {Stanford} Encyclopedia of Philosophy}. Metaphysics Research Lab, Stanford University, {W}inter 2025 edition.

\bibitem[Gurnee et~al., 2025]{gurnee2025when}
Gurnee, W., Ameisen, E., Kauvar, I., Tarng, J., Pearce, A., Olah, C., and Batson, J. (2025).
\newblock When models manipulate manifolds: The geometry of a counting task.
\newblock {\em Transformer Circuits Thread}.

\bibitem[Gurnee et~al., 2023]{gurnee_finding_2023}
Gurnee, W., Nanda, N., Pauly, M., Harvey, K., Troitskii, D., and Bertsimas, D. (2023).
\newblock Finding {Neurons} in a {Haystack}: {Case} {Studies} with {Sparse} {Probing}.
\newblock {\em Transactions on Machine Learning Research}.

\bibitem[Gurnee and Tegmark, 2024]{gurnee_language_2024}
Gurnee, W. and Tegmark, M. (2024).
\newblock Language {Models} {Represent} {Space} and {Time}.
\newblock \_eprint: 2310.02207.

\bibitem[Hanna et~al., 2023]{hanna_how_2023}
Hanna, M., Liu, O., and Variengien, A. (2023).
\newblock How does {GPT}-2 compute greater-than?: {Interpreting} mathematical abilities in a pre-trained language model.
\newblock In {\em Thirty-seventh {Conference} on {Neural} {Information} {Processing} {Systems}}.

\bibitem[Harman, 1987]{harman1987nonsolipsistic}
Harman, G. (1987).
\newblock (non-solipsistic) conceptual role semantics.
\newblock In Lepore, E., editor, {\em New Directions in Semantics}, pages 55--81. Academic Press, London.

\bibitem[Heap et~al., 2025]{heap_sparse_2025}
Heap, T., Lawson, T., Farnik, L., and Aitchison, L. (2025).
\newblock Sparse {Autoencoders} {Can} {Interpret} {Randomly} {Initialized} {Transformers}.
\newblock \_eprint: 2501.17727.

\bibitem[Henighan, 2024]{henighan_caloric_2024}
Henighan, T. (2024).
\newblock Caloric and the {Utility} of {Incorrect} {Theories}.

\bibitem[Herrmann and Levinstein, 2025]{herrmann2025}
Herrmann, D.~A. and Levinstein, B.~A. (2025).
\newblock Standards for belief representations in llms.
\newblock {\em Minds and Machines}, 35(1):1–25.

\bibitem[Hills, 2015]{hills2015understanding}
Hills, A. (2015).
\newblock Understanding why.
\newblock {\em Nous}, 50(4):661--688.

\bibitem[Hlobil, 2015]{hlobil2015antinormativism}
Hlobil, U. (2015).
\newblock Anti-normativism evaluated.
\newblock {\em International Journal of Philosophical Studies}, 23(3):376--395.

\bibitem[Jolicoeur-Martineau, 2025]{jolicoeur-martineau2025less}
Jolicoeur-Martineau, A. (2025).
\newblock Less is more: Recursive reasoning with tiny networks.
\newblock {\em arXiv preprint}, arXiv:2510.04871.
\newblock Preprint.

\bibitem[Kahneman, 2011]{kahneman2011thinking}
Kahneman, D. (2011).
\newblock {\em Thinking, Fast and Slow}.
\newblock Farrar, Straus and Giroux, New York, NY.

\bibitem[Kambhampati, 2025]{kambhampati2025memory}
Kambhampati, S. (2025).
\newblock Solving from memory vs solving from scratch --- or the futility of applying ``complexity lens'' to llms.
\newblock LinkedIn post. Available at \url{https://www.linkedin.com/posts/subbarao-kambhampati-3260708_sundayharangue-neurips2025-neurips2025-activity-7401246689677848578-XISo/}.

\bibitem[Kantamneni and Tegmark, 2025]{kantamneni_language_2025}
Kantamneni, S. and Tegmark, M. (2025).
\newblock Language {Models} {Use} {Trigonometry} to {Do} {Addition}.
\newblock \_eprint: 2502.00873.

\bibitem[Karvonen and Marks, 2025]{karvonen2025robustlyimprovingllmfairness}
Karvonen, A. and Marks, S. (2025).
\newblock Robustly improving llm fairness in realistic settings via interpretability.

\bibitem[Kelp, 2015]{Kelp2015}
Kelp, C. (2015).
\newblock Understanding phenomena.
\newblock {\em Synthese}, 192:3799--3816.
\newblock Received 19 August 2013; Accepted 12 November 2014; Published 07 January 2015; Issue date December 2015.

\bibitem[Khalifa, 2017]{khalifa2017scientific}
Khalifa, K. (2017).
\newblock {\em Understanding, explanation, and scientific knowledge}.
\newblock Cambridge University Press.

\bibitem[Kim et~al., 2025]{kim2025reasoningcircuitslanguagemodels}
Kim, G., Valentino, M., and Freitas, A. (2025).
\newblock Reasoning circuits in language models: A mechanistic interpretation of syllogistic inference.

\bibitem[Kitcher, 1981]{kitcher1981unification}
Kitcher, P. (1981).
\newblock Explanatory unification.
\newblock {\em Philosophy of science}, 48(4):507--531.

\bibitem[Kvanvig, 2003]{kvanvig2003value}
Kvanvig, J.~L. (2003).
\newblock {\em The value of knowledge and the pursuit of understanding}.
\newblock Cambridge University Press.

\bibitem[Kvanvig, 2009]{kvanvig2009value}
Kvanvig, J.~L. (2009).
\newblock The value of understanding.
\newblock In {\em Epistemic value}, pages 95--111.

\bibitem[Kvanvig, 2018]{kvanvig2018knowledge}
Kvanvig, J.~L. (2018).
\newblock Knowledge, understanding, and reasons for belief.
\newblock In Starr, D., editor, {\em The Oxford Handbook of Reasons and Normativity}, pages 685--705. Oxford University Press.

\bibitem[Leask et~al., 2025]{leask_sparse_2025}
Leask, P., Bussmann, B., Pearce, M.~T., Bloom, J.~I., Tigges, C., Moubayed, N.~A., Sharkey, L., and Nanda, N. (2025).
\newblock Sparse {Autoencoders} {Do} {Not} {Find} {Canonical} {Units} of {Analysis}.
\newblock In {\em The {Thirteenth} {International} {Conference} on {Learning} {Representations}}.

\bibitem[Levinstein and Herrmann, 2024]{levinstein2024}
Levinstein, B.~A. and Herrmann, D.~A. (2024).
\newblock Still no lie detector for language models: Probing empirical and conceptual roadblocks.
\newblock {\em Philosophical Studies}, page 1–27.

\bibitem[Li et~al., 2023]{li_emergent_2023}
Li, K., Hopkins, A.~K., Bau, D., Viégas, F., Pfister, H., and Wattenberg, M. (2023).
\newblock Emergent {World} {Representations}: {Exploring} a {Sequence} {Model} {Trained} on a {Synthetic} {Task}.
\newblock In {\em The {Eleventh} {International} {Conference} on {Learning} {Representations}}.

\bibitem[Lieberum et~al., 2023]{lieberum_does_2023}
Lieberum, T., Rahtz, M., Kramár, J., Nanda, N., Irving, G., Shah, R., and Mikulik, V. (2023).
\newblock Does {Circuit} {Analysis} {Interpretability} {Scale}? {Evidence} from {Multiple} {Choice} {Capabilities} in {Chinchilla}.
\newblock \_eprint: 2307.09458.

\bibitem[Lin et~al., 2024]{lin_othellogpt_2024}
Lin, J., Schonbrun, J., Karvonen, A., and Rager, C. (2024).
\newblock {OthelloGPT} {Learned} a {Bag} of {Heuristics}.

\bibitem[Lindsey et~al., 2025]{lindsey_biology_2025}
Lindsey, J., Gurnee, W., Ameisen, E., Chen, B., Pearce, A., Turner, N.~L., Citro, C., Abrahams, D., Carter, S., Hosmer, B., Marcus, J., Sklar, M., Templeton, A., Bricken, T., McDougall, C., Cunningham, H., Henighan, T., Jermyn, A., Jones, A., Persic, A., Qi, Z., Thompson, T.~B., Zimmerman, S., Rivoire, K., Conerly, T., Olah, C., and Batson, J. (2025).
\newblock On the {Biology} of a {Large} {Language} {Model}.
\newblock {\em Transformer Circuits Thread}.

\bibitem[Liu et~al., 2022a]{liu2022towards}
Liu, Z., Kitouni, O., Nolte, N.~S., Michaud, E., Tegmark, M., and Williams, M. (2022a).
\newblock Towards understanding grokking: An effective theory of representation learning.
\newblock {\em Advances in Neural Information Processing Systems}, 35:34651--34663.

\bibitem[Liu et~al., 2022b]{liu2022omnigrok}
Liu, Z., Michaud, E.~J., and Tegmark, M. (2022b).
\newblock Omnigrok: Grokking beyond algorithmic data.
\newblock {\em arXiv preprint arXiv:2210.01117}.

\bibitem[Ma et~al., 2025]{ma2025memorization}
Ma, B., Li, R., Wang, Y., Tan, H., and Li, X. (2025).
\newblock Memorization $\neq$ understanding: Do large language models have the ability of scenario cognition?
\newblock In {\em Proceedings of the 2025 Conference on Empirical Methods in Natural Language Processing}, pages 20758--20774, Suzhou, China. Association for Computational Linguistics.

\bibitem[Makelov et~al., 2024]{makelov_towards_2024}
Makelov, A., Lange, G., and Nanda, N. (2024).
\newblock Towards {Principled} {Evaluations} of {Sparse} {Autoencoders} for {Interpretability} and {Control}.
\newblock In {\em {ICLR} 2024 {Workshop} on {Secure} and {Trustworthy} {Large} {Language} {Models}}.

\bibitem[McDougall, 2023]{mcdougall_induction_2023}
McDougall, C. (2023).
\newblock Induction heads - illustrated.

\bibitem[Merrill et~al., 2021]{merrill2021provable}
Merrill, W., Goldberg, Y., Schwartz, R., and Smith, N.~A. (2021).
\newblock Provable limitations of acquiring meaning from ungrounded form: What will future language models understand?
\newblock {\em Transactions of the Association for Computational Linguistics}, 9:1047--1060.

\bibitem[Merullo et~al., 2025]{merullo2025memorizationreasoningspectrumloss}
Merullo, J., Vatsavaya, S., Bushnaq, L., and Lewis, O. (2025).
\newblock From memorization to reasoning in the spectrum of loss curvature.

\bibitem[Mikolov et~al., 2013]{mikolov_linguistic_2013}
Mikolov, T., Yih, W.-t., and Zweig, G. (2013).
\newblock Linguistic {Regularities} in {Continuous} {Space} {Word} {Representations}.
\newblock In Vanderwende, L., Daumé~III, H., and Kirchhoff, K., editors, {\em Proceedings of the 2013 {Conference} of the {North} {American} {Chapter} of the {Association} for {Computational} {Linguistics}: {Human} {Language} {Technologies}}, pages 746--751, Atlanta, Georgia. Association for Computational Linguistics.

\bibitem[Milli{\`e}re and Buckner, 2024]{milliere2024philosophical}
Milli{\`e}re, R. and Buckner, C. (2024).
\newblock A philosophical introduction to language models--part i: Continuity with classic debates.
\newblock {\em arXiv e-prints}, pages arXiv--2401.

\bibitem[Millikan, 1989]{millikan1989}
Millikan, R.~G. (1989).
\newblock In defense of proper functions.
\newblock {\em Philosophy of Science}, 56(2):288–302.

\bibitem[Millikan, 2017]{Millikan2017-MILBCU}
Millikan, R.~G. (2017).
\newblock {\em Beyond Concepts: Unicepts, Language, and Natural Information}.
\newblock OUP, Oxford, United Kingdom.

\bibitem[Minder et~al., 2025]{minder_latent_2025}
Minder, J., Dumas, C., Chughtai, B., and Nanda, N. (2025).
\newblock Latent {Scaling} {Robustly} {Identifies} {Chat}-{Specific} {Latents} in {Crosscoders}.
\newblock In {\em Sparsity in {LLMs} ({SLLM}): {Deep} {Dive} into {Mixture} of {Experts}, {Quantization}, {Hardware}, and {Inference}}.

\bibitem[Mitchell, 2025]{mitchell_llms_2025}
Mitchell, M. (2025).
\newblock {LLMs} and {World} {Models}, {Part} 2: {Evidence} {For} (and {Against}) {Emergent} {World} {Models} in {LLMs}.
\newblock {\em AI: A Guide for Thinking Humans}.

\bibitem[Mitchell and Krakauer, 2023]{mitchellkrakauer2023}
Mitchell, M. and Krakauer, D.~C. (2023).
\newblock The debate over understanding in ai’s large language models.
\newblock {\em Proceedings of the National Academy of Sciences}, 120(13):1–5.

\bibitem[Mollo and Millière, 2025]{mollo2025vectorgroundingproblem}
Mollo, D.~C. and Millière, R. (2025).
\newblock The vector grounding problem.

\bibitem[Mondorf and Plank, 2024]{mondorf2024beyond}
Mondorf, P. and Plank, B. (2024).
\newblock Beyond accuracy: Evaluating the reasoning behavior of large language models--a survey.
\newblock {\em arXiv preprint arXiv:2404.01869}.

\bibitem[Müller and Löhr, 2026]{muellerloehr}
Müller, V.~C. and Löhr, G. (2026).
\newblock {\em Artificial Minds}.
\newblock Cambridge University Press, Cambridge.

\bibitem[Nanda et~al., 2023a]{nanda_modular_addition_2023}
Nanda, N., Chan, L., Lieberum, T., Smith, J., and Steinhardt, J. (2023a).
\newblock Progress measures for grokking via mechanistic interpretability.
\newblock In {\em The {Eleventh} {International} {Conference} on {Learning} {Representations}}.

\bibitem[Nanda et~al., 2023b]{nanda_othello_2023}
Nanda, N., Lee, A., and Wattenberg, M. (2023b).
\newblock Emergent {Linear} {Representations} in {World} {Models} of {Self}-{Supervised} {Sequence} {Models}.
\newblock In Belinkov, Y., Hao, S., Jumelet, J., Kim, N., McCarthy, A., and Mohebbi, H., editors, {\em Proceedings of the 6th {BlackboxNLP} {Workshop}: {Analyzing} and {Interpreting} {Neural} {Networks} for {NLP}}, pages 16--30, Singapore. Association for Computational Linguistics.

\bibitem[Nanda et~al., 2023c]{nanda_fact_2023}
Nanda, N., Rajamanoharan, S., Kramar, J., and Shah, R. (2023c).
\newblock Fact finding: {Attempting} to reverse-engineer factual recall on the neuron level.
\newblock In {\em Alignment {Forum}}, page~6.

\bibitem[Nash, 1963]{nash1963}
Nash, L. (1963).
\newblock {\em The Nature of the Natural Sciences}.
\newblock Little, Brown, Boston.

\bibitem[Neander, 2017]{neander2017}
Neander, K. (2017).
\newblock {\em A Mark of the Mental: In Defense of Informational Teleosemantics}.
\newblock MIT Press, Cambridge, MA.

\bibitem[Nikankin et~al., 2025]{nikankin_arithmetic_2025}
Nikankin, Y., Reusch, A., Mueller, A., and Belinkov, Y. (2025).
\newblock Arithmetic {Without} {Algorithms}: {Language} {Models} {Solve} {Math} with a {Bag} of {Heuristics}.
\newblock In {\em The {Thirteenth} {International} {Conference} on {Learning} {Representations}}.

\bibitem[{Nostalgebraist}, 2024a]{nostalgebraist_information_2024}
{Nostalgebraist} (2024a).
\newblock Information {Flow} in {Transformers}.

\bibitem[{Nostalgebraist}, 2024b]{nostalgebraist_interpreting_2024}
{Nostalgebraist} (2024b).
\newblock Interpreting {GPT}: {The} {Logit} {Lens}.

\bibitem[Olah et~al., 2020]{olah2020zoom}
Olah, C., Cammarata, N., Schubert, L., Goh, G., Petrov, M., and Carter, S. (2020).
\newblock Zoom in: An introduction to circuits.
\newblock {\em Distill}.
\newblock https://distill.pub/2020/circuits/zoom-in.

\bibitem[Olsson et~al., 2022]{olsson_-context_2022}
Olsson, C., Elhage, N., Nanda, N., Joseph, N., DasSarma, N., Henighan, T., Mann, B., Askell, A., Bai, Y., Chen, A., Conerly, T., Drain, D., Ganguli, D., Hatfield-Dodds, Z., Hernandez, D., Johnston, S., Jones, A., Kernion, J., Lovitt, L., Ndousse, K., Amodei, D., Brown, T., Clark, J., Kaplan, J., McCandlish, S., and Olah, C. (2022).
\newblock In-context {Learning} and {Induction} {Heads}.
\newblock {\em Transformer Circuits Thread}.

\bibitem[{OpenAI}, 2025]{openai2025gpt5systemcard}
{OpenAI} (2025).
\newblock {GPT-5} system card.
\newblock Technical report, OpenAI.

\bibitem[OpenAI et~al., 2024]{openai2024gpt4technicalreport}
OpenAI, Achiam, J., Adler, S., Agarwal, S., Ahmad, L., ..., Zhuk, W., and Zoph, B. (2024).
\newblock {GPT-4} technical report.

\bibitem[Palod et~al., 2025]{palod2025performativethinkingbrittlecorrelation}
Palod, V., Valmeekam, K., Stechly, K., and Kambhampati, S. (2025).
\newblock Performative thinking? the brittle correlation between cot length and problem complexity.

\bibitem[Peacocke, 1992]{peacocke1992study}
Peacocke, C. (1992).
\newblock {\em A Study of Concepts}.
\newblock MIT Press, Cambridge, MA.

\bibitem[Piantadosi and Hill, 2022]{piantadosi2022meaning}
Piantadosi, S. and Hill, F. (2022).
\newblock Meaning without reference in large language models.
\newblock In {\em NeurIPS 2022 Workshop on Neuro Causal and Symbolic AI (nCSI)}.

\bibitem[Piantadosi et~al., 2024]{piantadosi2024concepts}
Piantadosi, S.~T., Muller, D.~C., Rule, J.~S., Kaushik, K., Gorenstein, M., Leib, E.~R., and Sanford, E. (2024).
\newblock Why concepts are (probably) vectors.
\newblock {\em Trends in Cognitive Sciences}, 28(9):844--856.

\bibitem[Power, 2022]{power2022conf}
Power, A. (2022).
\newblock If at first you don’t succeed, try try again.
\newblock In {\em Girl Geek X OpenAI}.

\bibitem[Power et~al., 2022]{power2022grokking}
Power, A., Burda, Y., Edwards, H., Babuschkin, I., and Misra, V. (2022).
\newblock Grokking: Generalization beyond overfitting on small algorithmic datasets.
\newblock {\em arXiv preprint arXiv:2201.02177}.

\bibitem[Pritchard, 2014]{pritchard2014}
Pritchard, D. (2014).
\newblock {\em Knowledge and Understanding}.
\newblock Synthese Library. Springer, United Kingdom.

\bibitem[Queloz, 2025a]{queloz2025a}
Queloz, M. (2025a).
\newblock Can ai rely on the systematicity of truth? the challenge of modelling normative domains.
\newblock {\em Philosophy \& Technology}, 38(34):1–27.

\bibitem[Queloz, 2025b]{queloz2025b}
Queloz, M. (2025b).
\newblock Explainability through systematicity: The hard systematicity challenge for artificial intelligence.
\newblock {\em Minds and Machines}, 35(35):1–39.

\bibitem[Queloz and Beckmann, 2026]{QuelozManuscript-QUEWWC-2}
Queloz, M. and Beckmann, P. (2026).
\newblock Why we care about understanding: Competence through predictive compression.

\bibitem[Radford et~al., 2016]{radford2016unsupervisedrepresentationlearningdeep}
Radford, A., Metz, L., and Chintala, S. (2016).
\newblock Unsupervised representation learning with deep convolutional generative adversarial networks.

\bibitem[Radford et~al., 2019]{radford2019language}
Radford, A., Wu, J., Child, R., Luan, D., Amodei, D., and Sutskever, I. (2019).
\newblock Language models are unsupervised multitask learners.
\newblock {\em OpenAI Blog}, 1(8):9.

\bibitem[Riggs, 2003]{riggs2003balancing}
Riggs, W.~D. (2003).
\newblock Balancing our epistemic goals.
\newblock {\em No{\^u}s}, 37(2):342--352.

\bibitem[Sanderson, 2024]{sanderson2024llms}
Sanderson, G. (2024).
\newblock How might llms store facts | deep learning chapter 7.

\bibitem[Shanahan, 2024a]{shanahantalking2}
Shanahan, M. (2024a).
\newblock Still “talking about large language models”: Some clarifications.
\newblock {\em arXiv preprint arXiv:2412.10291}.

\bibitem[Shanahan, 2024b]{shanahantalking1}
Shanahan, M. (2024b).
\newblock Talking about large language models.
\newblock {\em Communications of the ACM}, 67(2):68–79.

\bibitem[{Shanahan, Murray}, 2024]{shanahan_murray_simulacra_2024}
{Shanahan, Murray} (2024).
\newblock Simulacra as conscious exotica.
\newblock {\em Inquiry}, 0(0):1--29.
\newblock Publisher: Routledge.

\bibitem[Sharkey et~al., 2025]{sharkey_open_2025}
Sharkey, L., Chughtai, B., Batson, J., Lindsey, J., Wu, J., Bushnaq, L., Goldowsky-Dill, N., Heimersheim, S., Ortega, A., Bloom, J., Biderman, S., Garriga-Alonso, A., Conmy, A., Nanda, N., Rumbelow, J., Wattenberg, M., Schoots, N., Miller, J., Michaud, E.~J., Casper, S., Tegmark, M., Saunders, W., Bau, D., Todd, E., Geiger, A., Geva, M., Hoogland, J., Murfet, D., and McGrath, T. (2025).
\newblock Open {Problems} in {Mechanistic} {Interpretability}.
\newblock \_eprint: 2501.16496.

\bibitem[Shea, 2021]{shea_moving_2021}
Shea, N. (2021).
\newblock Moving {Beyond} {Content}-{Specific} {Computation} in {Artificial} {Neural} {Networks}.
\newblock {\em Mind and Language}, 38(1):156--177.
\newblock Publisher: Blackwell.

\bibitem[Sodium, 2024]{sodium2024heuristics}
Sodium (2024).
\newblock (maybe) a bag of heuristics is all there is \& a bag of heuristics is all you need.
\newblock LessWrong.

\bibitem[Stoehr et~al., 2024]{stoehr2024localizing}
Stoehr, N., Gordon, M., Zhang, C., and Lewis, O. (2024).
\newblock Localizing paragraph memorization in language models.
\newblock In {\em Proceedings of an archival peer–reviewed conference/workshop (published as CoRR abs/2403.19851)}.

\bibitem[Strevens, 2013]{Strevens2013-STRNUW}
Strevens, M. (2013).
\newblock No understanding without explanation.
\newblock {\em Studies in History and Philosophy of Science Part A}, 44(3):510--515.

\bibitem[Su et~al., 2025]{su2025conceptscomponentsconceptagnosticattention}
Su, J., Kempe, J., and Ullrich, K. (2025).
\newblock From concepts to components: Concept-agnostic attention module discovery in transformers.

\bibitem[Sullivan, 2022]{sullivan2022}
Sullivan, E. (2022).
\newblock Understanding from machine learning models.
\newblock {\em The British Journal for the Philosophy of Science}, 73(1):109--133.

\bibitem[Summerfield, 2025]{summerfield2025these}
Summerfield, C. (2025).
\newblock {\em These Strange New Minds: How AI Learned to Talk and What It Means}.
\newblock Penguin Publishing Group, New York, NY.

\bibitem[Svete et~al., 2024]{svete2024}
Svete, A., Borenstein, N., Zhou, M., Augenstein, I., and Cotterell, R. (2024).
\newblock Can transformers learn $ n $-gram language models?
\newblock {\em arXiv preprint arXiv:2410.03001}.

\bibitem[Tarng et~al., 2025]{tarng2025visual}
Tarng, J., Goel, P., and Kauvar, I. (2025).
\newblock Visual features across modalities: {SVG} and {ASCII} art reveal cross-modal understanding.
\newblock Circuits Updates -- October 2025. Edited by Joshua Batson and Adam Jermyn.

\bibitem[Templeton et~al., 2024]{templeton_scaling_2024}
Templeton, A., Conerly, T., Marcus, J., Lindsey, J., Bricken, T., Chen, B., Pearce, A., Citro, C., Ameisen, E., Jones, A., Cunningham, H., Turner, N.~L., McDougall, C., MacDiarmid, M., Freeman, C.~D., Sumers, T.~R., Rees, E., Batson, J., Jermyn, A., Carter, S., Olah, C., and Henighan, T. (2024).
\newblock Scaling {Monosemanticity}: {Extracting} {Interpretable} {Features} from {Claude} 3 {Sonnet}.
\newblock {\em Transformer Circuits Thread}.

\bibitem[Tigges et~al., 2024]{tigges_language_2024}
Tigges, C., Hollinsworth, O.~J., Geiger, A., and Nanda, N. (2024).
\newblock Language {Models} {Linearly} {Represent} {Sentiment}.
\newblock In Belinkov, Y., Kim, N., Jumelet, J., Mohebbi, H., Mueller, A., and Chen, H., editors, {\em Proceedings of the 7th {BlackboxNLP} {Workshop}: {Analyzing} and {Interpreting} {Neural} {Networks} for {NLP}}, pages 58--87, Miami, Florida, US. Association for Computational Linguistics.

\bibitem[Valentino et~al., 2025]{valentino2025mitigatingcontenteffectsreasoning}
Valentino, M., Kim, G., Dalal, D., Zhao, Z., and Freitas, A. (2025).
\newblock Mitigating content effects on reasoning in language models through fine-grained activation steering.

\bibitem[Valmeekam et~al., 2025]{valmeekam2025semanticsunreasonableeffectivenessreasonless}
Valmeekam, K., Stechly, K., Palod, V., Gundawar, A., and Kambhampati, S. (2025).
\newblock Beyond semantics: The unreasonable effectiveness of reasonless intermediate tokens.

\bibitem[Variengien, 2023]{variengien_common_2023}
Variengien, A. (2023).
\newblock Some {Common} {Confusion} {About} {Induction} {Heads}.
\newblock {\em LessWrong}.

\bibitem[Varma et~al., 2023]{varma2023explaining}
Varma, V., Shah, R., Kenton, Z., Kram{\'a}r, J., and Kumar, R. (2023).
\newblock Explaining grokking through circuit efficiency.
\newblock {\em arXiv preprint arXiv:2309.02390}.

\bibitem[Voita et~al., 2024]{voita-etal-2024}
Voita, E., Ferrando, J., and Nalmpantis, C. (2024).
\newblock Neurons in large language models: Dead, n-gram, positional.
\newblock In Ku, L.-W., Martins, A., and Srikumar, V., editors, {\em Findings of the Association for Computational Linguistics: ACL 2024}, pages 1288--1301, Bangkok, Thailand. Association for Computational Linguistics.

\bibitem[Wang et~al., 2023]{wang_interpretability_2023}
Wang, K.~R., Variengien, A., Conmy, A., Shlegeris, B., and Steinhardt, J. (2023).
\newblock Interpretability in the {Wild}: a {Circuit} for {Indirect} {Object} {Identification} in {GPT}-2 {Small}.
\newblock In {\em The {Eleventh} {International} {Conference} on {Learning} {Representations}}.

\bibitem[Wilkenfeld, 2018]{wilkenfeld2018compression}
Wilkenfeld, D.~A. (2018).
\newblock Understanding as compression.
\newblock {\em Philosophical Studies}, 176(10):2807--2833.

\bibitem[Williams, 2026]{WilliamsForthcoming-WILCSC-4}
Williams, I. (2026).
\newblock Can structural correspondences ground real world representational content in large language models?
\newblock {\em Mind and Language}.

\bibitem[Williams et~al., 2025]{williams2025mechanisticinterpretabilityneedsphilosophy}
Williams, I., Oldenburg, N., Dhar, R., Hatherley, J., Fierro, C., Rajcic, N., Schiller, S.~R., Stamatiou, F., and Søgaard, A. (2025).
\newblock Mechanistic interpretability needs philosophy.

\bibitem[Wittgenstein, 1953]{Wittgenstein1953-WITPI-4}
Wittgenstein, L. (1953).
\newblock {\em Philosophical Investigations}.
\newblock Wiley-Blackwell, New York, NY, USA.

\bibitem[Wright, 1973]{wright1973}
Wright, L. (1973).
\newblock Functions.
\newblock {\em Philosophical Review}, 82(2):139–168.

\bibitem[Yetman, 2025]{yetman2025representationlargelanguagemodels}
Yetman, C.~C. (2025).
\newblock Representation in large language models.

\bibitem[Yuan and Søgaard, 2025]{yuan_revisiting_2025}
Yuan, Y. and Søgaard, A. (2025).
\newblock Revisiting the {Othello} {World} {Model} {Hypothesis}.
\newblock \_eprint: 2503.04421.

\bibitem[Zhou et~al., 2024]{zhou_pre-trained_2024}
Zhou, T., Fu, D., Sharan, V., and Jia, R. (2024).
\newblock Pre-trained {Large} {Language} {Models} {Use} {Fourier} {Features} to {Compute} {Addition}.
\newblock In {\em The {Thirty}-eighth {Annual} {Conference} on {Neural} {Information} {Processing} {Systems}}.

\end{thebibliography}

\end{document}